\documentclass[runningheads]{llncs}

\usepackage{eccv} 

\usepackage{eccvabbrv}

\usepackage{graphicx}
\usepackage{booktabs}

\usepackage[accsupp]{axessibility}  

\usepackage{hyperref}

\usepackage{orcidlink}
\usepackage{multirow}
\usepackage{colortbl}

\usepackage{tcolorbox}
\usepackage{wrapfig}
\usepackage{soul}
\usepackage[utf8]{inputenc}
\usepackage{algorithm}
\usepackage{algpseudocode}
\usepackage{amsmath}
\usepackage{amsfonts}
\usepackage{color, xcolor}
\usepackage{fontawesome}
\usepackage{bbding}
\tcbuselibrary{skins, breakable, theorems}
\begin{document}

\title{Can Multimodal Large Language Models Truly Understand Small Objects?} 

\titlerunning{SOUBench}

\author{Fujun Han\inst{1,2} \and 
Junan Chen\inst{2}\and
Xintong Zhu\inst{2}\and
Jingqi Ye\inst{1,3}\and
Xuanjie Mao\inst{4}\and \\
Tao Chen\inst{4}\and
Peng Ye\inst{1,5}\thanks{Corresponding Author}
}

\authorrunning{Han et al.}

\institute{$^{1}$~Shanghai AI Laboratory,
$^{2}$~The Chinese University of Hong Kong, Shenzhen \\
$^{3}$~University of Science and Technology of China,
$^{4}$~Fudan University\\
$^{5}$~MMLab, The Chinese University of Hong Kong \\
\email{hanfujun@cuhk.edu.cn}}




\maketitle

\newtcbtheorem[auto counter]{cmt1}{Preliminary System Prompt}{
	colbacktitle = black!60!white, colframe = black!60!white,
	colback = black!5!white,
	fonttitle=\bfseries,
    fontupper=\itshape,
    float*,
    width=\textwidth,      
    boxrule=0.1pt,         
    left=4mm, right=4mm,  
    sharp corners,         
}{t}

\newtcbtheorem[auto counter]{cmt2}{Supervised Fine-Tuning Dataset Template}{
	colbacktitle = black!60!white, colframe = black!60!white,
	colback = black!5!white,
	fonttitle=\bfseries,
    fontupper=\itshape,
    float*,
    width=\textwidth,      
    boxrule=0.1pt,         
    left=4mm, right=4mm,   
    sharp corners,         
}{t}

\begin{abstract}
  Multimodal Large Language Models (MLLMs) have shown promising potential in diverse understanding tasks, \textit{e.g.}, image and video analysis, math and physics olympiads.  
  However, they remain blank and unexplored for Small Object Understanding (SOU) tasks. To fill this gap, we introduce \textbf{SOUBench}, the first and comprehensive benchmark for exploring the small objects understanding capability of existing MLLMs. Specifically, we first design an effective and automatic visual question-answer generation strategy, constructing a new SOU-VQA evaluation dataset, with 18,204 VQA pairs, six relevant sub-tasks, and three dominant scenarios (\textit{i.e.}, Driving, Aerial, and Underwater). Then, we conduct a comprehensive evaluation on \textcolor{black}{15} state-of-the-art MLLMs and reveal their weak capabilities in small object understanding. Furthermore, we develop \textit{SOU-Train}, a multimodal training dataset with 11,226 VQA pairs, to improve the SOU capabilities of MLLMs. Through supervising fine-tuning of the latest MLLM, we demonstrate that SOU-Train can effectively enhance the latest MLLM's ability to understand small objects. Comprehensive experimental results demonstrate that, the proposed SOUBench, along with the SOU-VQA and SOU-Train datasets, provides a crucial empirical foundation to the community for further developing models with enhanced small object understanding capabilities. Datasets and Code: \textcolor{blue}{\url{https://github.com/Hanfj-X/SOU}}.
\end{abstract}

\section{Introduction}
\label{sec:intro}

Multimodal Large Language Models (MLLMs) have shown outstanding performance across multiple domains and attracted considerable attention, \textit{e.g.}, Science Reasoning~\cite{GSM8K, math, hipho}, Video Understanding~\cite{Mvbench, Favor-bench, Video-bench}, Image Perception~\cite{mmbench, MMStar, MMUBench}, Scene Understanding~\cite{mllm-isu}. Besides, with the rapid development of MLLMs, 
the research paradigms for different downstream tasks have also changed greatly, \textit{e.g.}, for detection tasks, the detection models based on MLLMs outperform traditional state-of-the-art detection models~\cite{lmmdet} in the COCO benchmark~\cite{coco}. In Video Anomaly Detection, the research paradigm is undergoing a significant shift from boundary learning in the visual space to understanding and reasoning in the semantic space~\cite{VAD}. These paradigm shifts and performance improvements yield exciting and promising achievements. \par 
\begin{wrapfigure}{r}{0.565\textwidth}
\centering
\includegraphics[width=0.55\textwidth]{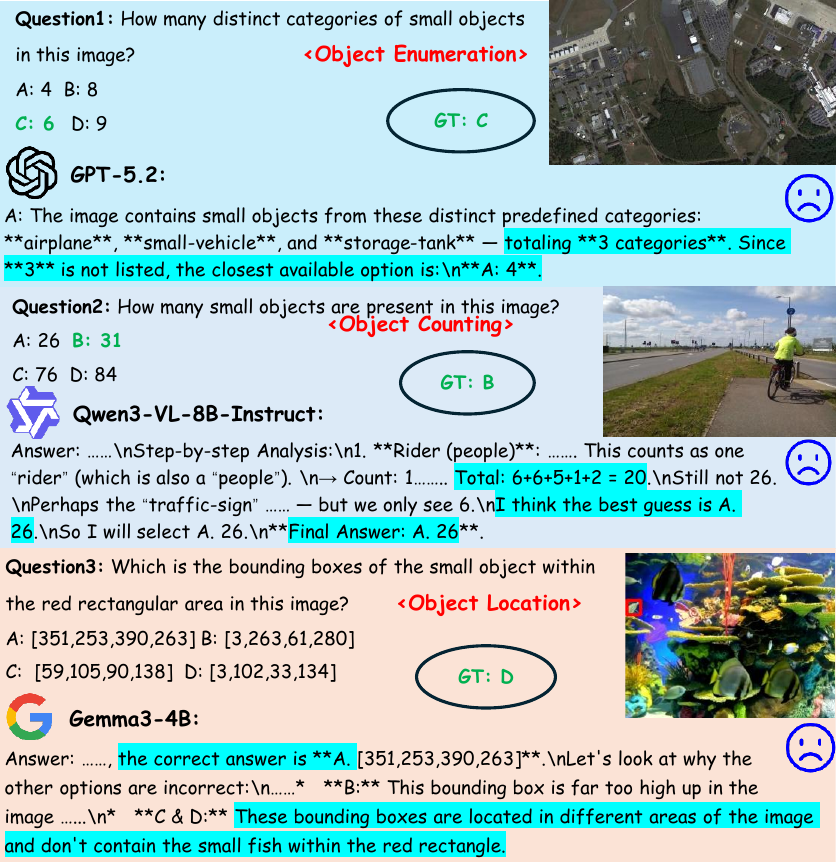}
\caption{Performance and comparison on different small object understanding tasks. We can find that, in several common small object understanding sub-tasks, the promising MLLMs provide incorrect answers. These phenomena indicate that the current state-of-the-art Multimodal Large Language Models still exhibit weak small object understanding capabilities.}
\label{gofdjhkfdjhkglkljhgfjlghkllfg}
\vspace{-6mm}
\end{wrapfigure}

Despite current MLLMs show strong capabilities and performance across various tasks, the task of small object understanding still remains unexplored and blank. Small Object Understanding (SOU), as one of the fundamental and crucial vision tasks for the MLLMs field, plays a crucial role in multiple application scenarios, \textit{e.g.}, autonomous driving~\cite{han2024mf}, remote sensing~\cite{Lhrs-bot, EarthGPT}, and underwater~\cite{RUOD}. Consequently, we have to raise basic yet important questions: \textit{How about the capability of current MLLMs in understanding small objects?} Based on these questions, we conduct some exploration experiments on state-of-the-art proprietary and open-source, as shown in Fig.~\ref{gofdjhkfdjhkglkljhgfjlghkllfg}. We can observe that even the most promising MLLMs, \textit{e.g.}, GPT-5.2~\cite{gpt-5} and Qwen3-VL-8B-Instruct~\cite{qwen3}, Gemma3-4B~\cite{team2025gemma} still have a limited understanding of small objects. For instance, for some simple understanding sub-tasks (\textit{e.g.}, small object Enumeration, Counting, and Location), MLLMs give incorrect analysis and report wrong answers, which suggests that current multimodal large models do not truly understand small objects. \par
To comprehensively and in-depth explore the capabilities of MLLMs on small object understanding tasks, we propose a first comprehensive benchmark for the small object understanding task, \textit{i.e.},  \textbf{S}mall \textbf{O}bject \textbf{U}nderstanding \textbf{B}enchmark (\textbf{SOUBench}). Specifically, we first design an automated and efficient \textbf{V}isual \textbf{Q}uestion \textbf{A}nswering (VQA) generation strategy to facilitate the acquisition of final sub-task VQA pairs. Simultaneously, we verify the effectiveness of our generated strategy on five different datasets with three dominant small object scenarios, \textit{i.e.}, Driving, Aerial, and Underwater. Finally, we generate a comprehensive SOU-VQA dataset with \textbf{18,204} VQA evaluation Pairs and six relevant sub-tasks, \textit{i.e.}, two Foundational Perception sub-tasks: Category Enumeration, Object Counting, two Spatial Reasoning sub-tasks: Category Recognition, Object Location, two Fine-Grained Understanding tasks: Specific Category Counting, Peripheral Object Identification. \par

\begin{figure}[t]
\centering
\includegraphics[scale=0.36]{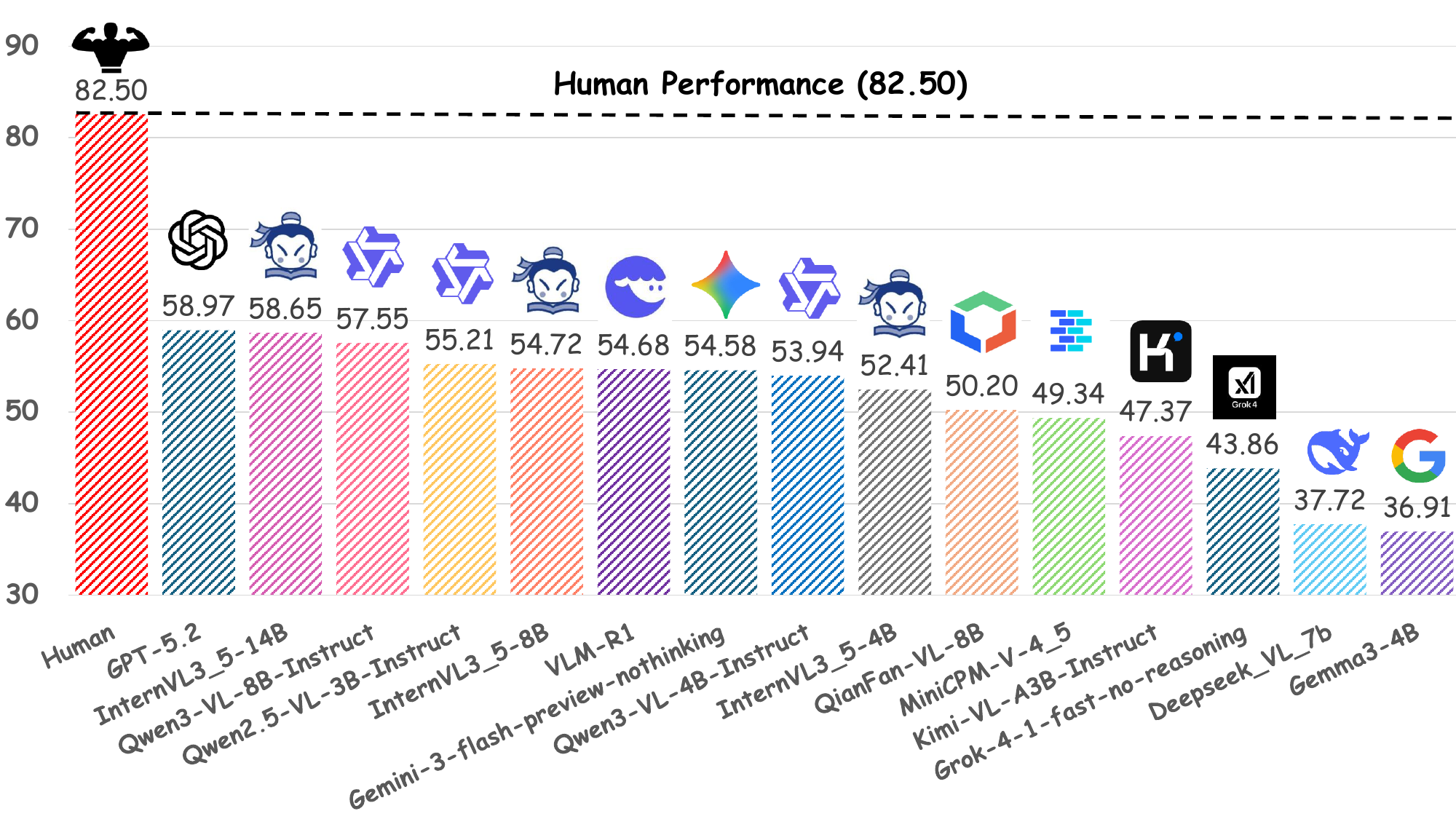}
\caption{The Performance and Comparison of 15 dominant MLLMs on the proposed small object understanding (SOU) task. We can find that the best performance is 58.97\% of GPT-5.2. However, compared to human performance (82.50\%), even the best GPT-5.2 still exhibits a significant gap,  which indicates that current MLLMs have weak capabilities in given SOU tasks.}
\label{fdashfjhasjdkfhajdsfhajksdhj}
\end{figure}

Comprehensive results on the SOU-VQA dataset reveal the key limitation of small object understanding tasks, as shown in Fig.~\ref{fdashfjhasjdkfhajdsfhajksdhj}. In state-of-the-art proprietary and open-source models, the best models are GPT-5.2 (58.97\%) and InternVL3\_5-14B (58.65\%), both of which are still below 60.00\%. Besides, compared to human performance, even the best-performing GPT-5.2 models still lag behind by 23.53\%. These performance gaps indicate that current Multimodal Large Language models exhibit weak capabilities in understanding small objects. To fill the gaps and improve the small object understanding capacity of current MLLMs, we also propose SOU-Train, a comprehensive supervised fine-tuning training dataset, with \textbf{11,226} rich and fine-grained VQA annotations. Note that our training VQA annotations are also automatically generated and can be extended to other object detection datasets. To validate the effectiveness of our obtained dataset and generated strategies, we fine-tune the current latest MLLMs, \textit{e.g.}, Qwen3-VL-2B-Instruct, based on our SOU-Trian. Sufficient experiment and comparison results demonstrate that our proposed SOU-Trian can achieve significant performance gains across different scenarios, \textit{e.g.}, +1.71\% (Driving), +1.17\% (Aerial), and +1.27\% (Underwater).

In summary, our contributions can be summarized as: \par
(1) 
To the best of our knowledge, MLLMs based Small Object Understanding (SOU) tasks 
are proposed for the first time. A comprehensive benchmark (\textbf{SOUBench}), including relative datasets and baselines, is reported for the specific task. SOUBench fully reveals the shortcomings of current MLLMs in understanding small objects. All datasets and codes will be publicly available. \par
(2) We design an effective automatic visual question-answer generation
pipeline and introduce a comprehensive SOU-VQA evaluation dataset for small object understanding tasks, with \textbf{18,204} pairs and six relevant sub-tasks. 
Comprehensive experiments and comparisons are conducted in 15 state-of-the-art MLLMs to evaluate the small object understanding capability of MLLMs. Sufficient results reveal that current MLLMs have a weak understanding ability in the proposed tasks, even the best MLLM is still behind Human performance by 23.53\%. \par
(3) We further construct SOU-Train, a multimodal VQA training dataset with \textbf{11,226} fine-grained annotations, to supervise the fine-tuning of the latest MLLM. The result denotes that the SOU-Train can effectively improve the small understanding ability of MLLM in different scenarios. Our research provides a crucial empirical foundation for the enhancement of the small object understanding capabilities of MLLMs.

\section{Related Work}
\textbf{Traditional Small Object Detection.} Small object detection is a specialized technique for identifying small-pixel objects in images, widely applied in surveillance~\cite{han2024ada}, autonomous driving~\cite{hanovid}, and remote image analysis~\cite{li2025hstnet}. Based on the historical development of small object detection tasks, this task can be divided into three main directions: Representation Enhancement, Perception Refinement, and Data \& Context Modeling, respectively~\cite{smallobject,chen2020dynamic,wu2022uiu}. However, although these works effectively address the problem of small object detection, they can only accomplish the single task of detection and not the multi-task of small object scenario understanding. Thus, in this paper, we propose a new and vital task to explore the problem, \textit{i.e.}, Multimodal Large Language Models based on small objects understanding. \par
\noindent \textbf{Multimodal Large Language Models Benchmarks.} Multimodal Large Language Models (MLLMs) have shown encouraging performance under various tasks due to their powerful reasoning and understanding abilities, \textit{e.g.}, Science (GSM8k~\cite{GSM8K}, MATH~\cite{math}, HiPho~\cite{hipho}), Image Understanding (ChartQA~\cite{Chartqa}, MMMU~\cite{MMMU}). Besides, in certain specialized fields, MLLMs have also exhibited great potential, \textit{e.g.}, General medical AI~\cite{Gmai}, Intrusion scene understanding~\cite{mllm-isu}, Object detection~\cite{lmmdet,ye2025dynamic}, Remote sensing~\cite{vrsbench, Lhrs-bot, EarthGPT, Xlrs-bench}. Despite the outstanding performance achieved by MLLMs in multiple fields, the capability of MLLMs in the domain of small object understanding remains unexplored. To address this limitation, we present a comprehensive benchmark for the small object understanding task, including SOU-VQA evaluation datasets, rich baselines, and an effective SOU-Train dataset.

\section{SOU-VQA Dataset}
\subsection{Data Collections}
We have thoroughly reviewed the development and progress in the small object field and find that the main small object task can be divided into three scenes, \textit{i.e.}, Driving, Aerial, and Underwater scenarios, respectively. These scenarios encompass multiple specialized small object datasets. For the driving scenario, we use the SODA-D~\cite{smallobject} due to its rich and diverse samples and categories.  
For the Aerial scenarios, we utilize three related datasets, VisDrone~\cite{VisDrone}, AI-TOD~\cite{AI-TOD}, and SODA-A~\cite{smallobject} to construct the VQA Pairs. The main reason is that the Aerial scenario has inherent complexity, covering both low-altitude and aerospace scenarios. For the Underwater scenarios, we adopt the RUOD~\cite{RUOD} dataset to make the VQA pairs. RUOD~\cite{RUOD} is a large Underwater dataset, with 10 categories, original images, and annotations. Note that there are no underwater datasets for small objects, so we adjust the original resolution of the RUOD dataset. In our SOU-VQA dataset, the maximum image resolution is 4731$\times$2789. Besides, for the definition of small objects, we adopt the default COCO~\cite{coco} format definition: \textit{An instance is classified as a small object if its absolute area is less than and equal to 1024 pixels}~\cite{smallobject}. The detailed information is shown in Tab.~\ref{mvkxbvnkcnbjkcvnbjnvcxjkbnxcjbnjcxn}. We can observe that SOU-VQA comprises 18204 VQA Pairs and 38 distinct small object categories, which provides a solid foundation for small object understanding tasks.

\begin{table*}[t]   
\begin{center}   
\caption{The source and statistics of the proposed SOU-VQA dataset. D, A, and U denote Driving, Aerial, and Underwater scenarios, respectively. $^\dagger$ denotes the size of the resolution after adjustment based on the small object understanding task requirement. $^\ast$ denotes that the dataset contains eight coordinates for the Object Location sub-task due to the characteristics of the dataset.}  
\label{mvkxbvnkcnbjkcvnbjnvcxjkbnxcjbnjcxn} 
\renewcommand\arraystretch{1.2} 
\scalebox{0.65} 
{
\begin{tabular}{ccccccc}   \hline
Datasets Source & Reference & Scenario  & Sampled Images & Image Resolution & VQA Pairs  &Small Object Categories \\ \hline   
SODA-D~\cite{smallobject} & TPAMI & 1 (Driving) & 1072 & 800$\times$600$\sim$3322$\times$1805 & 6432  & 9  \\   \hline 
VisDrone~\cite{VisDrone} & ICCV & 1 (Aerial) & 514 & 960$\times$540$\sim$1920$\times$1080 & 3084  & 10  \\
AI-TOD~\cite{AI-TOD} & ICPR & 1 (Aerial) & 410 & 800$\times$800 & 2460  & 8  \\
SODA-A~\cite{smallobject} & TPAMI & 1 (Aerial) & 328 & 1823$\times$1986$\sim$4731$\times$2789 & 1968$^\ast$  & 9 \\ \hline 
RUOD~\cite{RUOD} & Neurocomputing & 1 (Underwater) & 710 & 260$\times$233$\sim$1920$\times$1080$^\dagger$ & 4260  & 10  \\ \hline 
\rowcolor{gray!3}
SOU-VQA (Ours) & -& 3 (D\&A\&U) & 3034 & 260$\times$233$\sim$4731$\times$2789 & \textbf{18204} & 38 \\
 
\hline   
\end{tabular}
}
\end{center}   
\end{table*}


\begin{figure*}[t]
\centering
\includegraphics[scale=0.42]{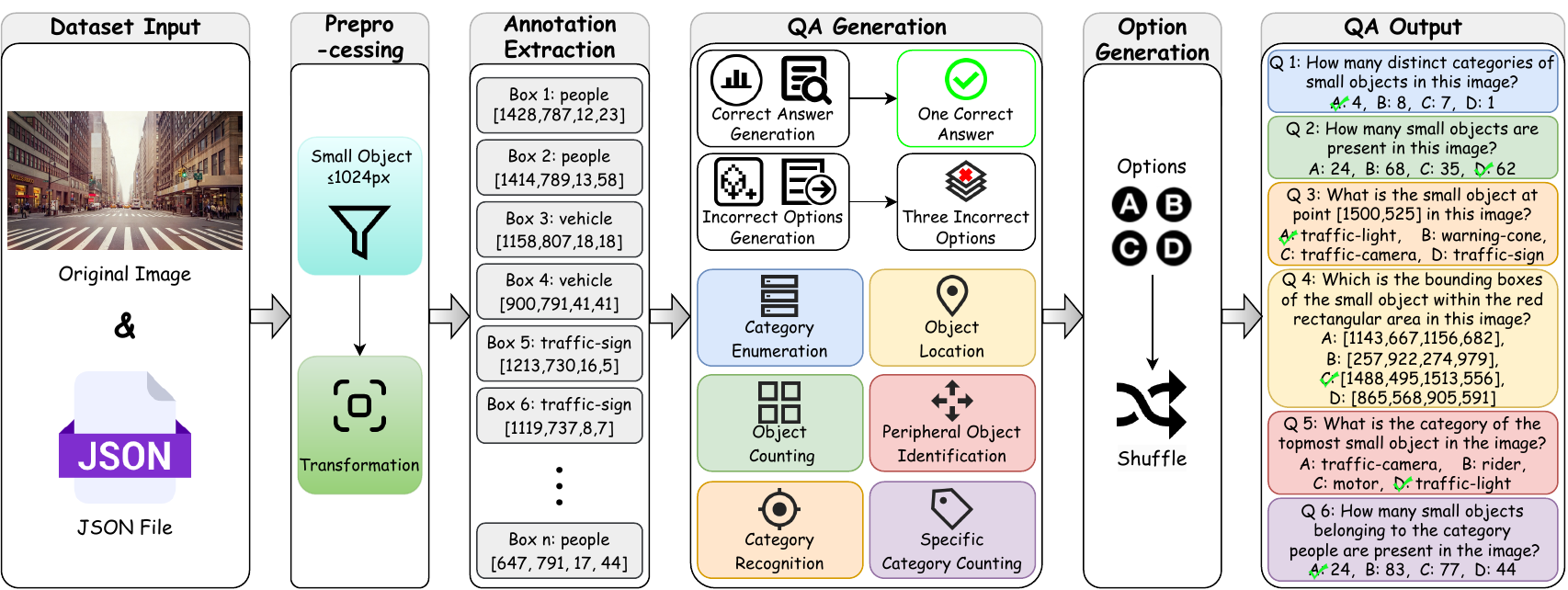}
\caption{The pipeline of automated generation strategy for Visual Question Answering (VQA). Our generation strategy primarily consists of six steps, \textit{i.e.}, Dataset Input, Preprocessing, Annotation Extraction, QA Generation, Option Generation, and QA output. In the QA Generation and Option Generation stage, we design automated procedures to obtain VQA pairs for six subtasks. Note that our strategy can be extended to other detection datasets to facilitate the generation of VQA pairs. Note that for underwater scenes, where relevant and comprehensive datasets are scarce. Therefore, we resized the original images to meet the requirements for constructing the Small Object Understanding VQA task. Besides, in SODA-A, datasets contain eight coordinates for location. Therefore, we also adopt this annotation method to construct our VQA pairs. Our detailed principles can be found in \textbf{Appendix A.}}
\label{iojicxjbkjkbjxkvxjc}
\end{figure*}
\subsection{Task Question Definition}
To comprehensively and systematically explore the capabilities of current MLLMs on small object understanding tasks, we carefully design six sub-tasks for this task from multiple views, \textit{i.e.}, two \textbf{Foundational Perception} sub-tasks: Category Enumeration, Object Counting, two \textbf{Spatial Reasoning} sub-tasks: Category Recognition, Object Location, two \textbf{Fine-Grained Understanding} tasks: Specific Category Counting, Peripheral Object Identification. The detailed description of each subtask is as follows: \par
\noindent $\rhd$  \textbf{Category Enumeration (CE).} This sub-task aims to evaluate whether the MLLMs can enumerate distinct semantic categories of small objects in the image. \par
\noindent $\rhd$ \textbf{Object Counting (OC).} This sub-task is to measure the MLLM's ability to accurately count the total number of small object instances in the image. \par
\noindent $\rhd$ \textbf{Category Recognition (CR).} This sub-task is to test a multimodal large model’s ability to recognize the semantic category of a specific small object indicated by a given point [x,y] in the image. Given an image and a spatial query in the form of pixel coordinates, the model is required to identify and name the category of the small object located at or nearest to the queried point. \par
\noindent $\rhd$ \textbf{Object Location (OL).} This sub-task is designed to estimate MLLM’s ability to localize small objects within a specified region of interest (ROI). Given an image and a red rectangular area, MLLMs are required to predict the bounding box(es) of the small object(s) contained within the region. \par
\noindent $\rhd$ \textbf{Specific Category Counting (SCC).}
This sub-task is designed to identify the quantity of given specific categories in an image, \textit{e.g.}, `people', `bicycle', `rider'. It evaluates the MLLM’s fine-grained understanding ability to count small objects that satisfy a given semantic category. \par
\noindent $\rhd$ \textbf{Peripheral Object Identification (POI).} This sub-task is to 
assess the MLLM’s fine-grained understanding ability to identify a small object by relative spatial position. Given an image and a peripheral spatial limitation, 
MLLMs must determine which small object satisfies the condition and report its category. \par

\subsection{VQA Pairs Generation Pipeline}

\begin{figure*}[t]
\centering
\includegraphics[scale=0.36]{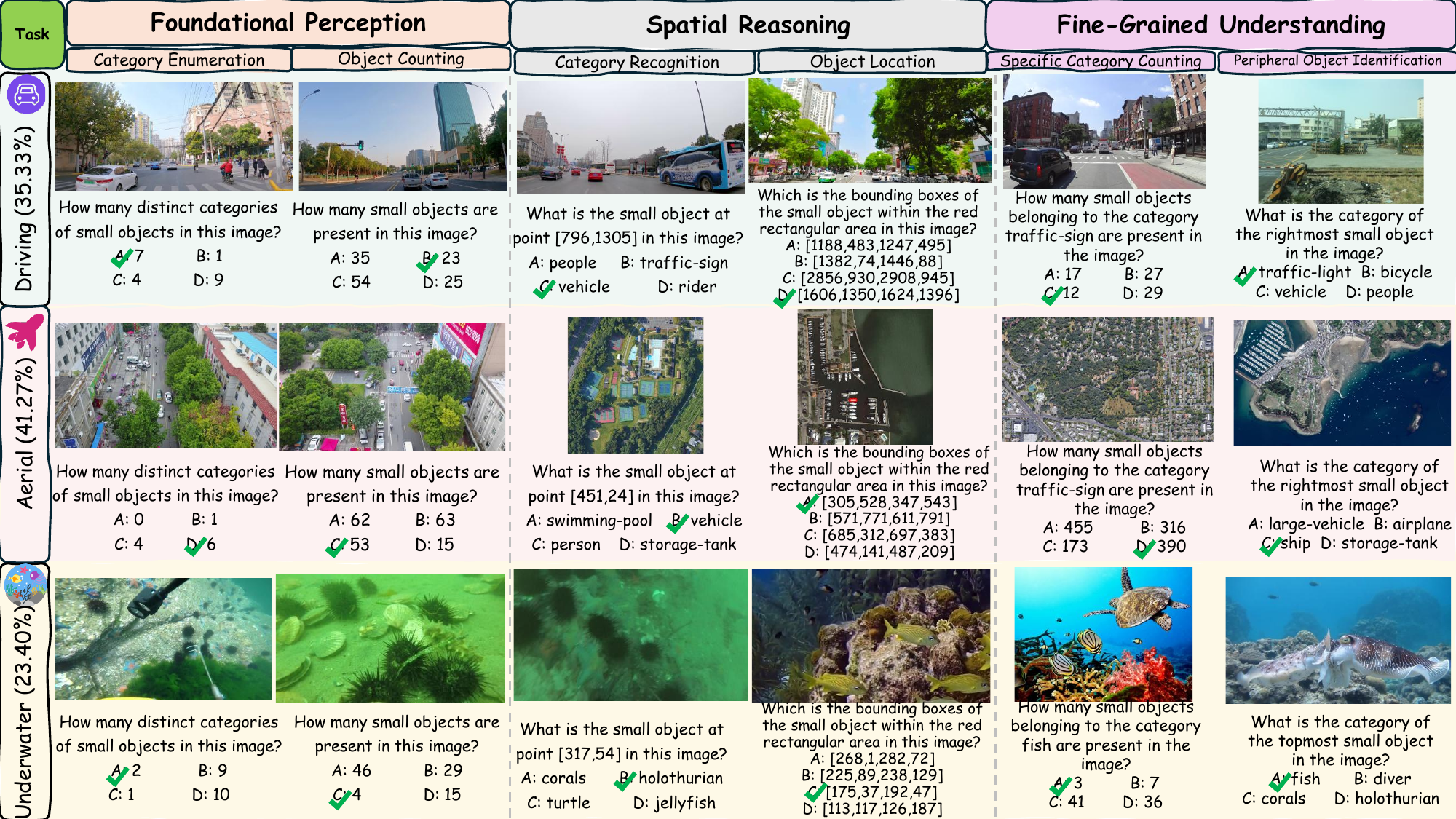}
\caption{An overview of the proposed SOU-VQA. Our VQA consists of three dominant small object scenarios: Driving, Aerial, and UnderWater. In each scenario, we design six distinct sub-tasks to comprehensively evaluate MLLMs' understanding ability of small objects, \textit{i.e.}, foundational Perception sub-tasks: Category Enumeration, Object Counting, Spatial Reasoning sub-tasks: Category Recognition, Object Location, Fine-Grained Understanding tasks: Specific Category Counting, Peripheral Object Identification. We can find that our SOU-VQA has rich and diverse scenarios and environments, which can enable a comprehensive evaluation of MLLMs' capabilities.}
\label{hkjfajdslkfjaklsdjfaklsdjkl}
\end{figure*}

\noindent \textbf{Automatic Question-Answer Generation.} In driving and aerial scenarios, the small object datasets are numerous and diverse. These specialized small object datasets provide a solid foundation for detection tasks in driving and aerial scenarios. However, in Underwater scenarios, the specialized small object datasets remain few. Therefore, we need to adjust the original images to meet the requirements for constructing the Small Object Understanding VQA task in the Underwater scenario. As illustrated in Fig.~\ref{iojicxjbkjkbjxkvxjc}, the generation pipeline comprises six sequential steps: Dataset Input, Preprocessing, Annotation Extraction, QA Generation, Option Generation, and QA Output, respectively. We first get the original datasets, including images and corresponding JSON annotation files. Then, to ensure the focus on small objects, input data undergoes a filtering process. Only objects with an area below the specific threshold of 1024 pixels ($\le 1024$px) are retained for our benchmark. Besides, to obtain the final small object datasets, for some original datasets, we adopt transformation techniques to resize images to meet the small object criteria. Thirdly, we automate the generation of the final VQA pairs through three sequential steps: 1) Annotation Extraction: The system parses the annotations to extract ground-truth data, specifically the category labels and bounding box coordinates for each identified small object. 2) QA Generation: This core module synthesizes question-answer pairs tailored to six specific tasks: Category Enumeration, Object Location, Object Counting, Peripheral Object Identification, Category Recognition, and Specific Category Counting. For each question, the system generates one ground-truth answer and three incorrect options (distractors). 3) Option Generation: The correct answer and distractors are aggregated and shuffled to assign random option labels (A, B, C, D), eliminating position bias. Finally, the generated text pairs are combined with the visual data to form the complete VQA instances. Note that our VQA pairs generation can extend other object detection datasets easily. Our detailed principles can be found in \textbf{Appendix A}. \par

\noindent \textbf{Manual verification.} To ensure the accuracy and quality of generation VQA pairs, a team of four students are formed to check the VQA pairs. Finally, our overall SOU-VQA dataset is presented in Fig.~\ref{hkjfajdslkfjaklsdjfaklsdjkl}. Our SOU-VQA contains a wide and diverse range of small object scenarios and can meet the requirements for a comprehensive evaluation of MLLM capabilities. \par

\subsection{Dataset Statistics and Comparisons}

\begin{figure}[t]
\centering
\includegraphics[width=0.92\linewidth]{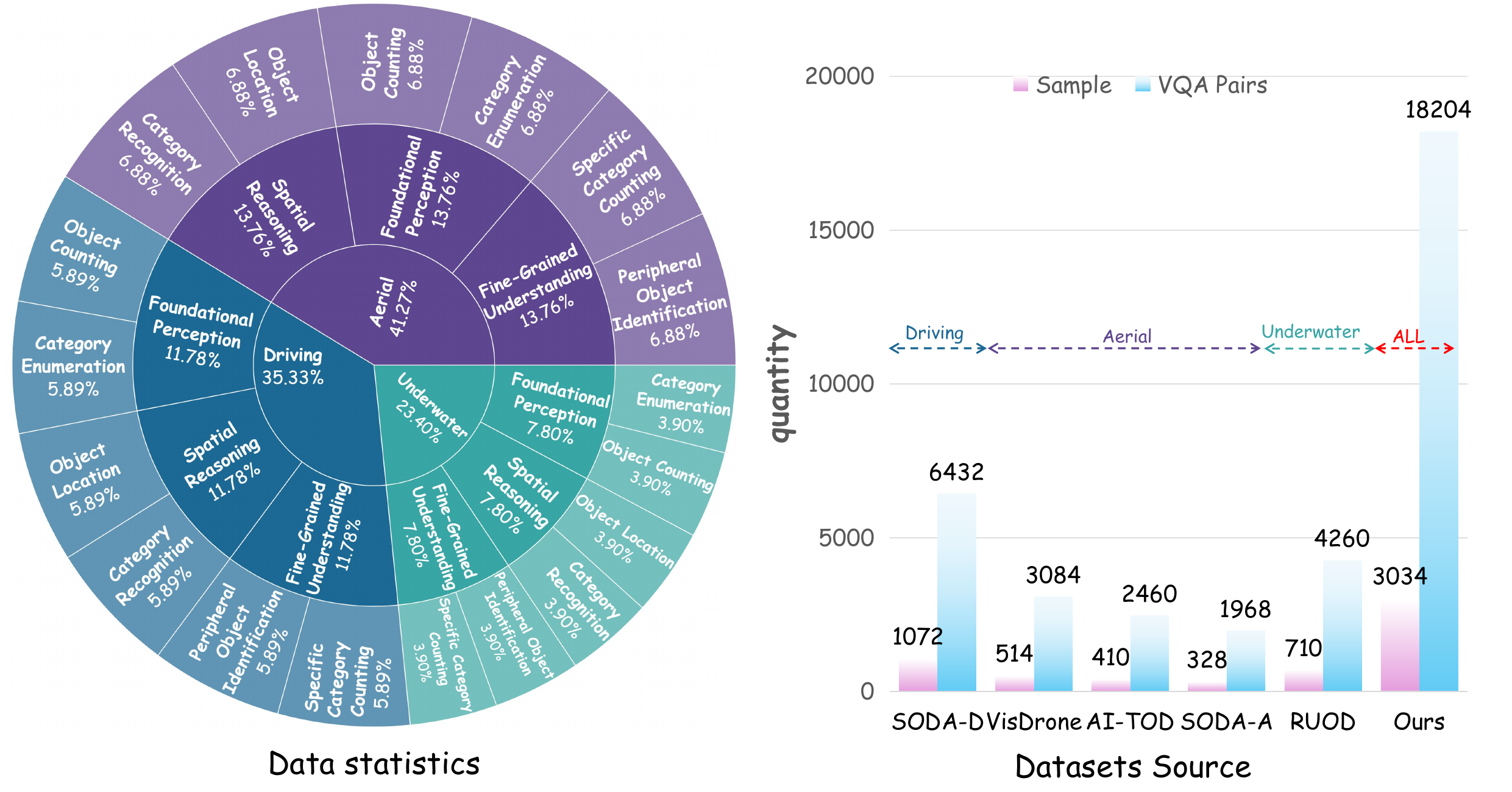}
\caption{The detailed dataset statistics. \textbf{Left}: Data statistics. \textbf{Right}: Datasets Source.}
\label{nmzvnlknngljdlgjdflskgjlkdfjk}
\end{figure}

\noindent \textbf{Dataset Statistics.} We then conduct a detailed statistical analysis on the proposed dataset, as shown in Fig.~\ref{nmzvnlknngljdlgjdflskgjlkdfjk}. We can observe that our SOU-VQA comprises a total of 18,024 VQA pairs across three primary scenarios: Driving (35.33\%), Aerial (41.27\%), and Underwater (23.40\%), respectively. In addition, each scenario contains six sub-tasks, accounting for Driving (5.89\%), Aerial (6.88\%), and Underwater (3.90\%) in total, respectively. Sufficient VQA and diverse scenarios provide a reliable data foundation for exploring the small object understanding capabilities of current MLLMs. \par

\noindent \textbf{Benchmark Comparisons.} To verify the advantages of the proposed SOUBench, we also compared it with other promising benchmarks, \textit{e.g.}, UWBench~\cite{zhang2025uwbench} and NuPlanQA~\cite{park2025nuplanqa}, etc. The comprehensive comparison is shown in Tab.~\ref{mkcxnbmkljsdkjhgkfjhkgdfkjfg}. We can find that the proposed SOUBench has multiple different perspectives, rich category diversity, and sufficient Eval QA. Besides, our SOUBench is specifically designed for small objects and can be used to assess the subtle capability, which demonstrates the advantages and feasibility of our SOUBench.

\begin{table*}[t]   
\begin{center}   
\caption{The comparison with other promising MLLM benchmarks. We compare some dominant benchmarks across six dimensions. D, A, U, and G denote the different scenarios: Driving, Aerial, Underwater, and General, respectively. \textcolor{red}{\faTimesCircle} indicates that this concept is not involved. \#Persp., \#CDS, \#SCA denotes Perspective, Category Diversity Scope, and Subtle Capability Assessment. Note that some datasets lack specific category counts, we manually determine them.} 
\label{mkcxnbmkljsdkjhgkfjhkgdfkjfg} 
\renewcommand\arraystretch{1.4} 
\scalebox{0.68}
{
\begin{tabular}{c|c|ccccccc}   \hline
Benchmark & \# Year & \# Object Pixels ($\le$ 1024)  & \# Persp.  & \# CDS. &  \# Scenes  & \# SCA & \# Eval QA  \\\hline 
UWBench~\cite{zhang2025uwbench} & Arxiv 2025 & \textcolor{red}{\faTimesCircle} & Single & Small (12) & U & \XSolidBrush & 124,983  \\ 
DisasterM3~\cite{wang2025disasterm3} & NeurIPS 2025 & Very few & Single & Small (9) & A & \XSolidBrush & 123,000  \\ 
CHOICE~\cite{an2024choice} & NeurIPS 2025 & Very few & Single & Middle (23) & A & \XSolidBrush & 8,700  \\ 
CoralVQA~\cite{han2025coralvqa} & NeurIPS 2025 & \textcolor{red}{\faTimesCircle} & Single & Small (1)  & U & \XSolidBrush & 277,653 & \\ 
MLLM-ISU~\cite{mllm-isu} & NeurIPS 2025 & Part & Single & Part (19)  & D & Part & 3,000 \\ 
XLRS-Bench~\cite{Xlrs-bench} & CVPR 2025 & Very few & Single & Middle (16)  & A & Very few & 45,942 \\ 
DriveBench~\cite{xie2025vlms} & CVPR 2025 & \textcolor{red}{\faTimesCircle} & Single & Middle (16)  & D & \XSolidBrush & 20,498  \\ 
3DSRBench~\cite{3dsrbench} & ICCV 2025 & \textcolor{red}{\faTimesCircle} & Two & Small (12)  & G & \XSolidBrush & 2,772 \\ 
VLADBench~\cite{li2025fine} & ICCV 2025 & \textcolor{red}{\faTimesCircle} & Single & Middle (29)  & D & \XSolidBrush & 12,000 \\ 
NuPlanQA~\cite{park2025nuplanqa} & ICCV 2025 & \textcolor{red}{\faTimesCircle} & Multiple & Small (9)  & D & \XSolidBrush &  8,000 \\ 
SOUBench (Ours) & - & All & Multiple & High (38)  & Multiple (D\&A\&U) & \Checkmark & 18,204 \\ 
\hline   
\end{tabular}
}
\end{center}   
\end{table*}

\section{Experiment and Results}
\subsection{Experiment setting}
\textbf{Comparison benchmarks models.} 
To comprehensively report the performance of current Multimodal Large Language models on small object understanding tasks, we conduct performance evaluations on 15 dominant state-of-the-art MLLMs, comprising 3 proprietary models (GPT-5.2~\cite{gpt-5}, Gemini-3-flash-preview-nothinking~\cite{gemini}, Grok-4-1-fast-no-reasoning~\cite{grok-4.1}) and 12 open-source models (Kimi-VL-A3B-Instruct~\cite{Kimi-VL}, Gemma3-4B~\cite{team2025gemma}, Qwen3-VL-4B-Instruct~\cite{qwen3}, Qwen3-VL-8B-Instruct~\cite{qwen3}, Qwen2.5-VL-3B-Instruct~\cite{Qwen2.5-VL}, InternVL3\_5-8B~\cite{Internvl3_5}, InternVL3\_5-14B~\cite{Internvl3_5}, QianFan-VL-8B~\cite{Qianfan-vl}, VLM-R1~\cite{VLM-R1}, MiniCPM-V-4\_5~\cite{MiniCPM-V4.5}, Deepseek\_VL\_7B~\cite{lu2024deepseekvl}, Deepseek-VL2-tiny~\cite{deepseekvl2}). For the proprietary models, we utilize APIs for testing. For open-source models, we use VLMEvalKit~\cite{vlmevalkit} for testing, and the configurations of the model is default parameters. Besides, we also report human performance metrics. Note that since different datasets and scenarios contain distinct predefined small object categories, we use different pre-prompts for definition and testing. The detailed illustration of the preliminary system prompt can be seen in \textbf{Appendix B}. In our experiments, we use \textbf{Accuracy} to report the final test performance.

\begin{table*}[h!]   
\begin{center}   
\caption{The Comprehensive evaluation results on 15 dominant MLLMs.\sethlcolor{red!20}\hl{CE}, \sethlcolor{red!20}\hl{OC} denotes the two sub-tasks of foundational Perception, \sethlcolor{blue!20}\hl{CR}, \sethlcolor{blue!20}\hl{OL} denotes the two sub-tasks of Spatial Reasoning, \sethlcolor{green!20}\hl{SCC}, \sethlcolor{green!20}\hl{POI} denotes the two sub-tasks of Fine-Grained Understanding. $^\dagger$ denotes the active parameters. \textbf{\underline{Bold}} and \underline{underlined} denotes the best and second-best model performance, respectively.} 

\renewcommand\arraystretch{0.89} 
\resizebox{\textwidth}{!}{

\begin{tabular}{l|c|c|cccccc|c}   \hline

Models & Scale & Scenarios  & \cellcolor{red!20} CE & \cellcolor{red!20} OC & \cellcolor{blue!20} CR & \cellcolor{blue!20} OL & \cellcolor{green!20} SCC & \cellcolor{green!20} POI & Average  \\  \hline

\rowcolor{red!1}
Random Chance  & - &   & 25.00 &  25.00 & 25.00 & 25.00& 25.00 & 25.00 & 25.00 \\ 
\rowcolor{red!1} 
Human Performance& - & & 76.19 & 76.19 &  90.48 & 100 & 80.95 & 85.71 & 84.92  \\ 
\rowcolor{red!1}
\rowcolor{red!1}
Gemini-3-flash-preview-nothinking~\cite{gemini}     &  -  &  &   \textbf{\underline{57.56}} &    32.09 &    58.86 &    30.32 &    64.27 &    \underline{65.00} & 51.35 \\
\rowcolor{red!1}
GPT-5.2~\cite{gpt-5}     &  -  &  &  44.78 &  61.94 &  61.10 &  46.08 &  \textbf{\underline{86.75}} &  \textbf{\underline{66.32}}  & 61.16 \\
\rowcolor{red!1}
Grok-4-1-fast-no-reasoning~\cite{grok-4.1}     & - &  &  31.62 &  54.85 & 43.66 &  26.03 &  75.19 &  50.00  & 46.89  \\  
\rowcolor{red!1}
\rowcolor{red!1}
Deepseek-VL2-tiny~\cite{deepseekvl2} & 3B/1B$^\dagger$ &  & 7.18 & 13.53 &  36.10 & 27.71 & 15.58 & 31.90 & 22.00  \\
\rowcolor{red!1}
InternVL3\_5-8B~\cite{Internvl3_5} & 8B & & 43.00 & 64.18 &    54.48 & 22.76 & 81.81 & 62.31 & 54.76\\
\rowcolor{red!1}
\rowcolor{red!1}
InternVL3\_5-14B~\cite{Internvl3_5}  & 14B &  & \underline{54.57} & \textbf{\underline{72.39}} &  60.35 & 25.00 & \underline{84.51} & 60.35 & 59.53 \\
\rowcolor{red!1}
Kimi-VL-A3B-Instruct~\cite{Kimi-VL}  & 16B/3B$^\dagger$ & Driving & 22.57 &  59.42 & 45.52 & 25.19 & 70.80 & 56.62 & 46.69\\
\rowcolor{red!1}
MiniCPM-V-4\_5~\cite{MiniCPM-V4.5}& 8B && 10.63 &    45.71 &    42.82 &    34.79 &    82.65 &    62.22 & 46.47 \\
\rowcolor{red!1}
QianFan-VL-8B ~\cite{Qianfan-vl}& 8B & & 20.80 &    \underline{71.08} &    42.72 &    24.91 &    83.12 &    54.01 & 49.44 \\
\rowcolor{red!1}
Qwen2.5-VL-3B-Instruct~\cite{Qwen2.5-VL} & 3B & & 45.06 &    61.01 &   \textbf{\underline{71.74}} & \underline{87.13} &    69.78 &    57.56 & \underline{65.38} \\
\rowcolor{red!1}
Qwen3-VL-4B-Instruct~\cite{qwen3} & 4B & & 32.84 & 62.59 &    39.46 & 27.33 &    81.06 &    59.98 & 50.56 \\
\rowcolor{red!1}
Qwen3-VL-8B-Instruct~\cite{qwen3} & 8B & & 43.10 &    65.67 & 49.53 & 24.91 &    81.72 &    61.10 & 54.34 \\
\rowcolor{red!1}
VLM-R1~\cite{VLM-R1} & 3B & & 42.07 &    65.49 &    \underline{69.96} &  \textbf{\underline{89.74}} &    72.20 &    54.10 & \textbf{\underline{65.59}} \\
\rowcolor{red!1}
Gemma3-4B~\cite{team2025gemma} & 4B & & 22.29 & 37.87 &    40.86 & 21.74 & 61.47 & 41.70  & 37.66 \\
\rowcolor{red!1}
Deepseek\_VL\_7B~\cite{lu2024deepseekvl} & 7B & &  6.16 &    33.21 &    50.28 &    22.29 &    68.38 &    51.49 & 38.64 \\
\hline  

\rowcolor{blue!1}
Random Chance  & - &   & 25.00 &  25.00 & 25.00 & 25.00 & 25.00 & 25.00 & 25.00 \\ 
\rowcolor{blue!1} 
Human Performance&- & & 68.00 & 68.00 &  88.00 & 100 & 84.00 & 88.00 & 82.67  \\ 
\rowcolor{blue!1}
\rowcolor{blue!1}
Gemini-3-flash-preview-nothinking~\cite{gemini}     &  -  &  &  \textbf{\underline{58.79}} &  38.34 &  67.97 & 43.45 &  47.60 & \textbf{\underline{73.64}} & 54.97 \\
\rowcolor{blue!1}
GPT-5.2~\cite{gpt-5}     &  -  &  &  41.05 &    35.86 &  \textbf{\underline{71.49}} &  \textbf{\underline{68.21}} &    53.91 &  \underline{70.69}  & \underline{56.87} \\
\rowcolor{blue!1}
Grok-4-1-fast-no-reasoning~\cite{grok-4.1}     & - &  & 27.80 &  27.24 & 51.60 & 37.94 &  41.13 &  46.49  & 38.70  \\  
\rowcolor{blue!1}
\rowcolor{blue!1}
Deepseek-VL2-tiny~\cite{deepseekvl2} & 3B/1B$^\dagger$ &  & 15.58 & 23.72 &  50.96 & 24.20 & 21.01 & 40.10 & 29.26  \\
\rowcolor{blue!1}
InternVL3\_5-8B~\cite{Internvl3_5} & 8B &  & 40.42 & \underline{38.66} & 64.54 & 58.47 & 54.55 & 62.62 & 53.21 \\
\rowcolor{blue!1}
InternVL3\_5-14B~\cite{Internvl3_5}  & 14B &  & \underline{56.31} & \textbf{\underline{41.69}} & 69.81 & 61.90 & \textbf{\underline{56.95}} & 64.62 & \textbf{\underline{58.55}}\\
\rowcolor{blue!1}
Kimi-VL-A3B-Instruct~\cite{Kimi-VL}  & 16B/3B$^\dagger$ & Aerial & 32.11 & 36.26 & 64.70 & 48.08 & 44.73 & 64.86 & 48.46\\
\rowcolor{blue!1}
MiniCPM-V-4\_5~\cite{MiniCPM-V4.5}& 8B & & 33.63 & 28.04 & 67.57 & \underline{68.13} & 48.08 & 63.02 & 51.41 \\
\rowcolor{blue!1}
QianFan-VL-8B ~\cite{Qianfan-vl}&8B & & 28.12 & 36.82 & 65.02 & 62.38 & \underline{54.63} & 62.54 & 51.58 \\
\rowcolor{blue!1}
Qwen2.5-VL-3B-Instruct~\cite{Qwen2.5-VL} &3B & & 47.20 & 29.31 & 68.21 & 57.83 & 42.73 & 58.95 & 50.71 \\
\rowcolor{blue!1}
Qwen3-VL-4B-Instruct~\cite{qwen3} & 4B& & 41.29 & 35.70 & 64.30 & 60.38 & 53.12 & 63.34 & 53.02 \\
\rowcolor{blue!1}
Qwen3-VL-8B-Instruct~\cite{qwen3} & 8B& & 52.48 & 36.74 & \underline{70.13} & 62.22 & 51.52 & 67.73 & 56.80 \\
\rowcolor{blue!1}
VLM-R1~\cite{VLM-R1} & 3B & & 46.17 & 30.27 & 67.09 & 58.87 & 42.73 & 57.27 & 50.40 \\
\rowcolor{blue!1}
Gemma3-4B~\cite{team2025gemma} & 4B & & 25.08 & 25.64 & 55.43 & 19.73 & 37.06 & 56.71 & 36.61 \\
\rowcolor{blue!1}
Deepseek\_VL\_7B~\cite{lu2024deepseekvl} & 7B & & 20.29 & 22.28 & 59.19 & 22.60 & 28.67 & 57.43 & 35.08 \\
\hline  

\rowcolor{green!1}
Random Chance  & - &   & 25.00 & 25.00 & 25.00 & 25.00 & 25.00 & 25.00 & 25.00 \\ 
\rowcolor{green!1} 
Human Performance& - & & 85.71 & 42.86 & 100 & 100 & 57.14 & 85.71 & 78.57  \\ 
\rowcolor{green!1}
\rowcolor{green!1}
Gemini-3-flash-preview-nothinking~\cite{gemini}     &  -  &  &   \textbf{\underline{77.32}} &    32.11 &    \underline{73.94} &    39.30 &    52.25 &    \textbf{\underline{77.61}} & 58.76 \\
\rowcolor{green!1}
GPT-5.2~\cite{gpt-5}     &  -  &  &  47.04 &    60.14 &    57.46 &    \textbf{\underline{68.31}} &    61.41 &    61.83  & 59.37 \\
\rowcolor{green!1}
Grok-4-1-fast-no-reasoning~\cite{grok-4.1}     & - &  & 37.75 &    49.15 &    60.85 & 29.01 &    57.89 &    55.63  & 48.38   \\  
\rowcolor{green!1}
Deepseek-VL2-tiny~\cite{deepseekvl2} & 3B/1B$^\dagger$ &  & 2.96 &    11.13 &    45.92 &    22.82 &    24.93 &    43.80 & 25.26  \\
\rowcolor{green!1}
InternVL3\_5-8B~\cite{Internvl3_5} & 8B & & 49.01 &    61.41 &    59.86 &    40.85 &    \textbf{\underline{77.18}} &    55.63 & 57.32 \\
\rowcolor{green!1}
InternVL3\_5-14B~\cite{Internvl3_5}  & 14B &  & 49.44 &    61.27 &    60.99 &    40.28 &    72.54 &    60.56 & 57.51 \\
\rowcolor{green!1}
Kimi-VL-A3B-Instruct~\cite{Kimi-VL}  & 16B/3B$^\dagger$ & Underwater & 22.96 &    39.30 &    61.13 &    41.13 &    56.20 &    58.17  & 46.48 \\
\rowcolor{green!1}
MiniCPM-V-4\_5~\cite{MiniCPM-V4.5}& 8B && 26.06 &    43.80 &    65.21 &    \underline{46.06} &    59.30 &    59.58  & 50.00 \\
\rowcolor{green!1}
QianFan-VL-8B ~\cite{Qianfan-vl}& 8B && 33.94 &    51.83 &    52.11 &    41.13 &    65.77 &    48.73 & 48.92 \\
\rowcolor{green!1}
Qwen2.5-VL-3B-Instruct~\cite{Qwen2.5-VL} & 3B& & 32.82 &    45.07 &    63.24 &    33.10 &    63.24 &    49.44 & 47.82 \\
\rowcolor{green!1}
Qwen3-VL-4B-Instruct~\cite{qwen3} & 4B& & 45.49 &    \textbf{\underline{66.06}} &    68.45 &    39.86 &    \underline{75.35} &    68.87 & \underline{60.68} \\
\rowcolor{green!1}
Qwen3-VL-8B-Instruct~\cite{qwen3} & 8B& & \underline{60.00} &    \underline{62.82} &    \textbf{\underline{74.23}} & 40.70 &   73.80 &    \underline{70.85} & \textbf{\underline{63.73}} \\
\rowcolor{green!1}
VLM-R1~\cite{VLM-R1} & 3B & & 38.45 &    41.27 &    56.48 &    33.94 &    58.03 &    46.34  & 45.75 \\
\rowcolor{green!1}
Gemma3-4B~\cite{team2025gemma} & 4B & & 14.23 & 19.30 & 56.48 & 22.82 & 49.72 & 55.35 & 36.32 \\
\rowcolor{green!1}
Deepseek\_VL\_7B~\cite{lu2024deepseekvl} & 7B & & 12.82 & 43.52 &    49.44 &    26.20 &    64.23 &    49.72 & 40.99\\
\hline 
\end{tabular}
}
\label{bvcbvcxbvcxbcv} 
\end{center}   
\end{table*}

\subsection{Main Results and Findings} \par
We conduct a comprehensive experiment and evaluation on the proposed SOU-VQA. The detailed
results are reported and shown in Tab.~\ref{bvcbvcxbvcxbcv} and Fig.~\ref{fdashfjhasjdkfhajdsfhajksdhj}. Note that the performance of Aerial is obtained by taking a comprehensive calculation and the average of the three aerial sub-datasets, \textit{i.e.}, A-VisDrone, A-AITOD, and A-SODA. Four main findings can be observed and summarized. \par
\noindent \textbf{Findings 1: How about the small object understanding capability of different dominant MLLMs?} We first report the understanding performance of the current dominant MLLMs. (1) From the Fig.~\ref{fdashfjhasjdkfhajdsfhajksdhj}, we can find that, in the tested MLLMs, 
The GPT-5.2~\cite{gpt-5} model achieves the best overall performance, reaching 58.97\%. The second performance model is the InternVL3\_5-14B~\cite{Internvl3_5}, with a performance of 58.65\%. (2) In different scenarios, MLLM models present varying capabilities in understanding small objects. In autonomous Driving scenarios, VLM-R1~\cite{VLM-R1} achieves the best performance at 65.59\%. In Aerial scenarios, InternVL3\_5-14B~\cite{Internvl3_5} reaches the best performance at 58.55\%. In Underwater scenarios, the best performance model is Qwen3-VL-8B-Instruct~\cite{qwen3}, with 63.73\%. (3) In three different scenarios, the rank of performance is Driving (65.59\%)$>$Underwater (63.73\%)$>$Aerial (58.55\%). We can observe that current MLLMs exhibit a limited understanding of Aerial scenarios. The main reason is that small object occupy very few pixels and lack sufficient visual detail, making it difficult for MLLMs to effectively align and fuse cross-modal features for accurate recognition. We think this will be a valuable direction for future research. \par

\noindent \textbf{Findings 2: Can current MLLMs really understand small objects?} From the Fig.~\ref{fdashfjhasjdkfhajdsfhajksdhj}, we can find that, although these models exhibit best performance in different scenarios, their overall performance remains low ($<$60\%). From Tab.~\ref{bvcbvcxbvcxbcv}, especially in certain tasks, \textit{e.g.}, Object Location sub-task in Driving, the performance of some MLLMs is even lower than 30\%. This indicates that current MLLMs still struggle with understanding small objects.\par 

\noindent \textbf{Findings 3: How far performance gap between the current MLLMs and humans on the small object understanding tasks?} To better evaluate the capability of current MLLMs in small objects understanding tasks, we compare the best-performing MLLM with human performance, as shown in Tab.~\ref{bvcbvcxbvcxbcv}. We can find that in three different small object scenarios, compared with human performance, the best MLLMs still lag behind human performance by \textbf{19.33\%} (Driving), \textbf{24.12\%} (Aerial), and \textbf{14.84\%} (Underwater), respectively. The performance gap denotes that the current MLLMs still have a long way to go in truly understanding small objects. \par

\noindent \textbf{Findings 4: Open-source Models vs Proprietary models.} We also compare Open-source models with Proprietary models. As the Fig.~\ref{fdashfjhasjdkfhajdsfhajksdhj}, the Proprietary model, \textit{e.g.}, GPT-5.2, still exhibits the best 58.97\% performance. However, we also find that in understanding small object tasks, the performance of some Open-source models, \textit{e.g.}, InternVL3\_5-14B~\cite{Internvl3_5} (58.65\%), Qwen3-VL-8B-Instruct~\cite{qwen3} (57.55\%), has gradually approached the Proprietary model.  \par

\subsection{Validation Experiments and Analyses}

\noindent \textbf{Training setting.} To better enhance the small object understanding capacity of MLLMs, we further propose \textbf{SOU-Train}, a multimodal training dataset to fine-tune current MLLMs. SOU-Train comprises 11,226 training VQA Pairs across three different scenarios. In these experiments, we adopt the LoRA~\cite{Lora} way to conduct supervised fine-tuning and set the lora\_rank as 8. Besides, due to the high resolution of the original image, we set image\_max\_pixels to 4194304 (\textit{i.e.}, resolution size: 2048$\times$2048). More settings can be found in \textbf{Appendix D}.\par
\begin{table*}[htb]   
\begin{center}   
\caption{The quantitative results on the proposed SOU-Train. $^\dagger$ denotes that the experiments are conducted in Aerial scenarios. We can find that in different scenarios, our proposed SOU-Train can effectively improve the understanding capacity of MLLM, which also proves the effectiveness of the SOU-Train.  }

\renewcommand\arraystretch{1.6} 
\scalebox{0.62}
{
\begin{tabular}{l|c|cccccc|c}   
\hline
Model & Scenario  & CE & OC & CR & OL & SCC & POI & Avg. \\  \hline
Qwen3-VL-2B-Instruct & \multirow{2}*{Driving}  &   37.22 &    54.57 &    32.93 & 17.91 &    70.15 &  41.79 & 42.43 \\
Qwen3-VL-2B-Instruct+SOU-Train & & 38.43\textcolor{gray}{$_{(+1.21)}$} &    54.20\textcolor{gray}{$_{(-0.37)}$} &    37.50\textcolor{gray}{$_{(+4.57)}$} &    18.38\textcolor{gray}{$_{(+0.47)}$} &    70.62\textcolor{gray}{$_{(+0.47)}$} &    45.71\textcolor{gray}{$_{(+3.92)}$} &    \textbf{44.14}\textcolor{black}{$_{(\textbf{+1.71})}$} \\ \hline

Qwen3-VL-2B-Instruct & \multirow{2}*{A-VisDrone$^\dagger$}  & 25.88 &    34.63 &    43.39 &    43.97 &    51.36 &    46.69  & 40.99 \\
Qwen3-VL-2B-Instruct+SOU-Train  & & 26.85\textcolor{gray}{$_{(+0.97)}$} &    32.88\textcolor{gray}{$_{(-1.75)}$} &    43.77\textcolor{gray}{$_{(+0.38)}$} &    47.67\textcolor{gray}{$_{(+3.70)}$} &    54.47\textcolor{gray}{$_{(+3.11)}$} &    46.50\textcolor{gray}{$_{(-0.19)}$} &    \textbf{42.02}\textcolor{black}{$_{(\textbf{+1.03})}$}  \\  \hline

Qwen3-VL-2B-Instruct & \multirow{2}*{A-AITOD$^\dagger$}  & 37.07 &    27.80 &    87.32 &    63.90 &    32.44 &    84.63  & 55.53 \\
Qwen3-VL-2B-Instruct+SOU-Train  & & 42.68\textcolor{gray}{$_{(+5.61)}$} &    25.85\textcolor{gray}{$_{(-1.95)}$} &    87.32\textcolor{gray}{$_{(+0.00)}$} &    65.61\textcolor{gray}{$_{(+1.71)}$} &    33.41\textcolor{gray}{$_{(+0.97)}$} &    82.93\textcolor{gray}{$_{(-1.70)}$} &    \textbf{56.30}\textcolor{black}{$_{(\textbf{+0.77})}$}  \\  \hline

Qwen3-VL-2B-Instruct & \multirow{2}*{A-SODA$^\dagger$}  & 27.13 &    28.35 &    72.87 &    23.48 &    32.93 &    67.99  & 42.12 \\
Qwen3-VL-2B-Instruct+SOU-Train & & 27.44\textcolor{gray}{$_{(+0.31)}$} &    33.23\textcolor{gray}{$_{(+4.88)}$} &    71.65\textcolor{gray}{$_{(-1.22)}$} &    26.22\textcolor{gray}{$_{(+2.74)}$} &    33.54\textcolor{gray}{$_{(+0.61)}$} &    67.68\textcolor{gray}{$_{(-0.31)}$} &    \textbf{43.29}\textcolor{black}{$_{(\textbf{+1.17})}$}  \\  \hline

Qwen3-VL-2B-Instruct & \multirow{2}*{Underwater}  &  45.35 & 48.59 &    67.89 &    30.14 & 62.39 &    66.34 & 53.45\\

Qwen3-VL-2B-Instruct+SOU-Train& &  48.03\textcolor{gray}{$_{(+2.68)}$} &    49.72\textcolor{gray}{$_{(+1.13)}$} &    69.58\textcolor{gray}{$_{(+1.69)}$} &    30.99\textcolor{gray}{$_{(+0.85)}$} &    63.66\textcolor{gray}{$_{(+1.27)}$} &    66.34\textcolor{gray}{$_{(+0.00)}$} &    \textbf{54.72}\textcolor{black}{$_{(\textbf{+1.27})}$}  \\
\hline
\end{tabular}
}
\label{bvcxbvcbxvcbxcvbcxbvcnbvnc} 
\end{center}   
\end{table*}

\noindent \textbf{Results Analyses.} Based on the proposed SOU-Train and above setting, we supervise fine-tuning the latest Qwen3-VL-2B-Instruct~\cite{qwen3} model and validated it in different scenarios and datasets, \textit{e.g.}, Driving, Underwater, and Aerial. Detailed results are shown in Tab.~\ref{bvcxbvcbxvcbxcvbcxbvcnbvnc}. We can find that, compared with the original Qwen3-VL-2B-Instruct model, our SOU-Train can effectively enhance model performance across distinct scenarios and datasets, \textit{i.e.}, 1.71\% (Driving), 1.03\% (A-VisDrone), 0.77\% (A-AITOD), 1.17\% (A-SODA), and 1.27\% (Underwater), respectively. These promising performance gains denote the effectiveness of the proposed SOU-Train dataset. \par

\subsection{More Exploration Experiments and Analyses}

\begin{figure*}[t]
\centering
\includegraphics[scale=0.35]{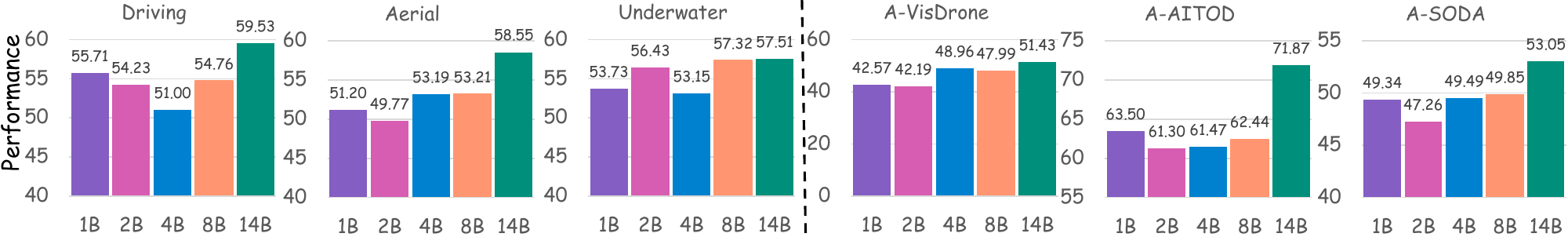}
\caption{The performance of five different model scales. Driving, Aerial, and Underwater denote the three different scenarios of the proposed SOU-VQA dataset. Note that the performance of Aerial is the overall performance of A-VisDrone, A-AITOD, and A-SODA, as shown in the three subplots on the right. We can observe that, in Driving, Aerial, and Underwater three scenarios, as the scale of the model increases, the understanding ability of small objects also basically grows, which suggests that increasing model capacity effectively mitigates the inherent challenges in SOU tasks.}
\label{poigoiosdgjijfdsigsjdfkgjsfggslkljkf}
\end{figure*}

\textbf{Experiments 1: To what extent does model scale influence small object understanding performance?} In this subsection, we explore the relationship between the small object understanding ability of MLLMs and model scale. After investigation, we find that the InternVL3\_5 series model have rich model scale. Thus, we conduct the detailed experiments based on the InternVL3\_5 series model, including 1B, 2B, 4B, 8B, and 14B, as shown in Fig. ~\ref{poigoiosdgjijfdsigsjdfkgjsfggslkljkf}. We can find that two findings are revealed: 1) In the three evaluated scenarios, model scale and fine-grained perception of small objects show a generally positive correlation, \textit{i.e.}, as the model size increases, the understanding ability of small objects gets better (55.71\%$\rightarrow$59.53\% in Driving, 51.20\%$\rightarrow$58.55\% in Aerial, 53.73\%$\rightarrow$57.51\% in Underwater). The consistent performance gain across diverse scenarios suggests that increasing model capacity effectively mitigates the inherent challenges in small object understanding. 2) Although uncommon, an interesting phenomenon is observed in certain scenarios, \textit{e.g.}, autonomous driving, the 4B model perform worse than the 1B model (51.00\% $<$ 55.71\%). We report the detailed results of all sub-tasks in Fig.~9 of the \textbf{Appendix E}. We can find that the 4B model have low understanding ability for Category Enumeration (19.12\%), which leads to the overall performance getting lower. These findings indicate that we require carefully balancing model scale with the specific demands to achieve better performance in different small objects understanding scenarios and tasks.
More detailed scale results can be found in \textbf{Appendix E}.\par

\begin{figure}[h]
\centering
\includegraphics[scale=0.36]{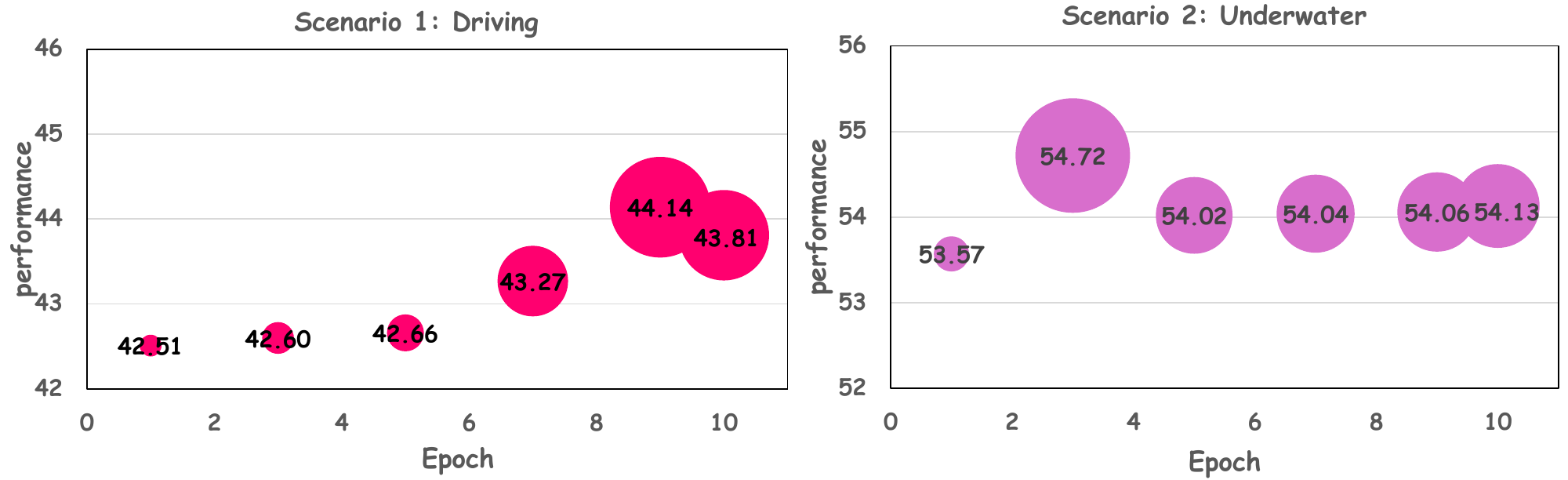}
\caption{The performance impact of different training costs.}
\label{ahgjkdfhgnxmblsjk}
\end{figure}

\noindent\textbf{Experiments 2: How significant is the impact of the training costs in different scenarios?}
In this experiment, we explore the impact of different training costs on small object understanding capabilities across various scenarios. We take the Driving and Underwater scenario as examples, as shown in Fig.~\ref{ahgjkdfhgnxmblsjk}. We can find that in different scenarios, the best performance can be reached at different training costs, \textit{i.e.}, Epoch 9 (Driving) and Epoch 3 (Underwater). We think the primary reason is that, for the MLLMs, the speed of learning small object features varies across different scenarios. In busy scenarios, \textit{e.g.}, city roads, the MLLM need more time to recognize small objects from a messy background. However, in simple scenes, \textit{e.g.}, underwater, the model learns the small object features much faster. Therefore, we suggest that researchers should not use the same number of epochs for every task. Specifically, for high-entropy scenarios with dense interference, a higher training cost is essential to capture fine-grained features, while for low-contrast or redundant environments, early stopping is recommended to prevent overfitting and ensure the best results. 

\vspace{-5pt}

\noindent \textbf{Experiments 3: Generalization Exploration.} 
In this experiment, we conduct an interesting generalization experiment. We attempt to directly apply the supervised fine-tuning model to other scenarios for testing, \textit{e.g.}, from Underwater scenarios to Driving scenarios. Specifically, we first perform SFT training in Underwater scenarios and then test the SFT model in Driving scenarios. 
\begin{wraptable}{r}{0.52\linewidth}
\centering
\vspace{-10mm}
\caption{The Performance results of cross-task transfer. U, D, and A denote Underwater, Driving, and Aerial scenarios. $^\dagger$ and $^\ddagger$ denote the proposed A-VisDrone and A-SODA in our experiments.}
\vspace{+9pt}
\renewcommand\arraystretch{1.02}
\resizebox{\linewidth}{!}{
\begin{tabular}{l|c|c|c}   \hline
Baseline Model & Transfer & Strategy &  Average \\  \hline
\multirow{5}*{Qwen3-VL-2B-Instruct~\cite{qwen3}} &  \XSolidBrush  & - & 42.43     \\ 
~ & \Checkmark & U$\rightarrow$D  & \textbf{43.00}\textcolor{black}{$_{(+0.57)}$}   \\ \cline{2-4}
& \XSolidBrush  & -  & 40.99 \\ 
~ & \Checkmark & U$\rightarrow$A$^\dagger$  & \textbf{41.47}\textcolor{black}{$_{(+0.48)}$}   \\ \hline

\multirow{5}*{Qwen3-VL-2B-Instruct~\cite{qwen3}} &  \XSolidBrush & - & 42.43     \\ 
~ & \Checkmark & A$^\ddagger$$\rightarrow$D  & \textbf{42.65}\textcolor{black}{$_{(+0.22)}$}   \\ \cline{2-4}
& \XSolidBrush  & - & 53.45 \\ 
~ & \Checkmark & A$^\ddagger$$\rightarrow$U  & \textbf{53.83}\textcolor{black}{$_{(+0.38)}$}   \\ \hline
\end{tabular}
}
\label{BNMCXDJGFDJHIGFJDIKHJIG}
\vspace{-6mm}
\end{wraptable}
To comprehensively evaluate the model's generalization ability, we conduct multiple experiments under different settings, as shown in Tab.~\ref{BNMCXDJGFDJHIGFJDIKHJIG}. We can find that models under supervision fine-tuned can effectively generalize to other unknown and untrained scenarios, \textit{e.g.}, in Underwater$\rightarrow$Driving. Compared with the original baseline, the final average performance surpasses it by 0.57\%. Similarly, some improvements are observed across other generalization tasks. The bidirectional performance gains across different scenarios reveal that our SOU tasks are a scale-sensitive rather than domain-specific task. Models capture a universal visual prior of geometry and local contrast that transcends environmental boundaries. Besides, these findings demonstrate that combining a basic sensitivity to small objects with the common sense of large models is feasible. More importantly, this suggests the possibility of building a universal model that understands small objects across multiple scenarios and environments, rather than training specialized models for each specific scenario.  multiple scenarios and environments, rather than training specialized models for each specific scenario.

\section{Conclusion}
In this paper, we propose SOUBench, the first and comprehensive benchmark to explore the small objects understanding capability of existing Multimodal Large Language Models. 
Specifically, we first propose an effective automatic VQA generation strategy and construct a new SOU-VQA dataset by automatic procedures, with 18,204 VQA evaluation Pairs, six relevant sub-tasks in three dominant scenarios, \textit{i.e.}, Driving, Aerial, and Underwater. Then, we conduct a comprehensive evaluation on 15 state-of-the-art MLLMs and reveal weak capabilities in small object understanding tasks. We further build SOU-Train, a multimodal VQA dataset with 11,226 VQA annotations, to supervise fine-tuning current MLLMs and improve the understanding capabilities of small objects. Comprehensive experiments demonstrate the effectiveness of our proposed dataset and strategy. Our proposed SOUBench, including generation strategy, SOU-VQA, SOU-Train, and rich baselines, provides a solid foundation for future improvements in MLLMs' understanding capability of small objects.


%
\bibliographystyle{splncs04}
\bibliography{main}

\clearpage


\noindent\textbf{APPENDIX OVERVIEW}\par
\vspace{+3pt}
\noindent\textbf{Table of contents:} \par
\vspace{+3pt}
\noindent $\bullet$ \textcolor{gray}{$\S$} \textbf{\textcolor{black}{A}: The detailed generation principles of VQA Pairs} \par
\vspace{+3pt}
\noindent $\bullet$ \textcolor{gray}{$\S$} \textbf{\textcolor{black}{B}: The details of preliminary system prompt}  \par
\vspace{+3pt}
\noindent $\bullet$ \textcolor{gray}{$\S$} \textbf{\textcolor{black}{C}: Evaluation of three sub-datasets in Aerial}  \par
\vspace{+3pt}
\noindent $\bullet$ \textcolor{gray}{$\S$} \textbf{\textcolor{black}{D}: The more detailed setting of training} \par
\vspace{+3pt}
\noindent $\bullet$ \textcolor{gray}{$\S$} \textbf{\textcolor{black}{E}: More results of different model scales} \par 

\vspace{+15pt}
\noindent \textbf{\large A: ~The detailed generation principles of VQA Pairs} \par
\begin{figure*}[h]
\centering
\includegraphics[scale=0.42]{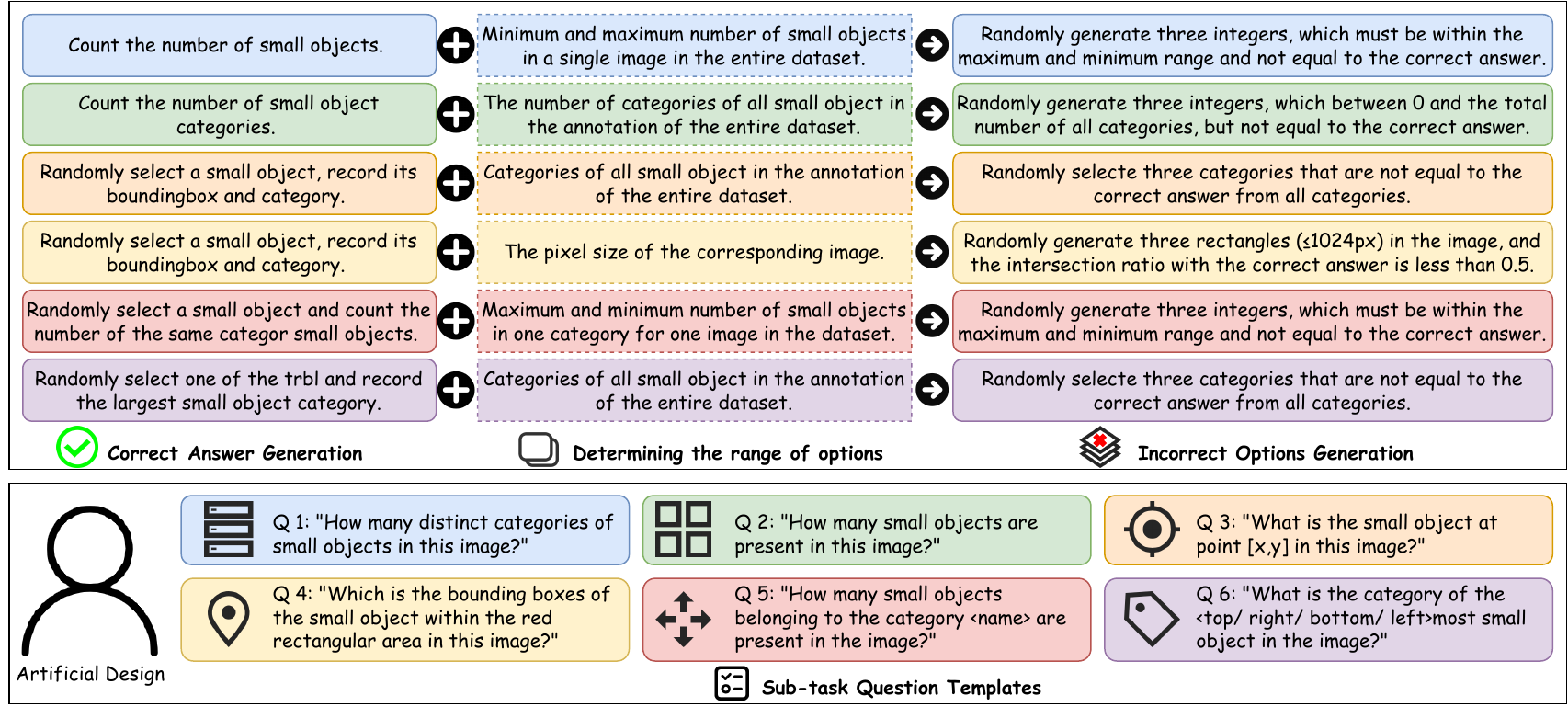}
\caption{The detailed principles of the proposed automated generation strategy of visual question answering. The principal mainly consists of three modules: Sub-task Question Template Design, Correct Answer Generation, and Incorrect Options Generation.}
\label{nmnbmcnkjxnjsdjfg}
\end{figure*}

\vspace{+15pt}
\noindent In this section, we provide detailed generation principles about our proposed generation strategies. The detailed architecture of the QA Generation module is illustrated in the Fig.~\ref{nmnbmcnkjxnjsdjfg}. QA Generation module primarily consists of three main parts: Sub-task Template Design, Correct Answer Generation, and Incorrect Options Generation, respectively. We explain every part as follows:
(1) Sub-task Question Template Design: We design six distinct question templates (Q1-Q6) focused on small object understanding, including Category Enumeration, Object Counting, Category Recognition, Object Location, Specific Category Counting, and Peripheral Object Identification. 
(2) Correct Answer Generation: Leveraging the ground-truth annotations, Correct answers are automatically generated by either counting specific small objects or randomly sampling the bounding boxes and categories of objects present in the image. 
(3) Incorrect Options Generation (Distractors): To formulate multiple-choice questions, we determine the range of potential options based on the statistical distribution of the entire dataset, such as the global minimum/maximum object counts and the complete category set. Three distractors are then randomly generated for each question, ensuring they fall within the valid range but do not overlap with the correct answer. 

\vspace{+15pt}
\noindent \textbf{\large B: ~The details of preliminary system prompt} \par
\vspace{+15pt}

\noindent In this paper, our SOU-VQA datasets are constructed based on multiple different datasets, \textit{e.g.}, SODA~\cite{smallobject}, VisDrone~\cite{VisDrone}, AI-TOD~\cite{AI-TOD}, and RUOD~\cite{RUOD}. Due to the diversity of dataset sources and the inconsistent definitions of small objects across each dataset, we need to design distinct pre-prompts for each dataset. As shown in Preliminary System Prompt 1, we provide two separate instructions, \textit{i.e.}, P1: the definition of small objects and P2: the definition of small object category under different source datasets.
Note that for the definition of small objects, we adopt the standard COCO dataset~\cite{coco}, \textit{i.e.}, An instance is classified as a small object if its absolute area is less than and equal to 1024 pixels~\cite{smallobject}. These two instructions (P1 and P2) effectively resolve category ambiguities between different datasets and better evaluate model performance. Besides, for the \textit{Predefined Small Object Category List}, these datasets span multiple distinct categories. Therefore, when testing, please select the corresponding \textit{Predefined Small Object Category List} based on the dataset name. Please note that \sethlcolor{yellow}\hl{yellow}, \sethlcolor{lime}\hl{lime}, and \sethlcolor{pink}\hl{pink} denote the Driving, Aerial, and Underwater scenarios, respectively.
\label{jkgjfkgjkfdsjgkldfjkgljfghkfl}
\begin{cmt1}{}{}
System prompt = 
You are an assistant specializing in the small objects scene understanding task. Please read the following question carefully and provide a correct option of the question with strictly adhering to the following instructions: \par
1. Small Object Definition: An instance is classified as a small object if its absolute area is less than and equal to 1024 pixels. \par
\sethlcolor{yellow}\hl{2. Predefined Small Object Category List: [`people', `rider', `bicycle', `motor', `vehicle', `traffic-sign', `traffic-light', `traffic-camera', `warning-cone'].} $\gets$If test `Driving' scenario, please notice it! \par
\sethlcolor{lime}\hl{2. Predefined Small Object Category List: [`airplane', `helicopter', `small-vehicle', `large-vehicle', `ship', `container', `storage-tank', `swimming-pool', `windmill'].} $\gets$If test `Aerial (SODA)' scenario, please notice it!  \par
\sethlcolor{lime}\hl{2. Predefined Small Object Category List: [`airplane', `bridge', `storage-tank', `ship', `swimming-pool', `vehicle', `person', `wind-mill'].} $\gets$If test `Aerial (AITOD)' scenario, please notice it! \par
\sethlcolor{lime}\hl{2. Predefined Small Object Category List: [`pedestrian', `person', `bicycle', `car', `van', `truck', `tricycle', `awning-tricycle', `bus', `motor'].} $\gets$If test `Aerial (VisDrone)' scenario, please notice it! \par
\sethlcolor{pink}\hl{2. Predefined Small Object Category List: [`holothurian', `echinus', `scallop', `starfish', `fish', `corals', `diver', `cuttlefish', `turtle', `jellyfish'].} $\gets$If test `Underwater' scenario, please notice it! \par
\end{cmt1}


\vspace{+15pt}
\noindent \textbf{\large C: ~Evaluation of three sub-datasets in Aerial} \par
\vspace{+15pt}

\noindent In this section, we evaluate and report the performance of three sub-datasets in the Aerial scenario, \textit{i.e.,} A-VisDrone, A-AITOD, and A-SODA, as shown in Tab.~\ref{BVCXBVCBXCVBXCVBXBXCVBXVCBVCB}. We can find that: (1) \textbf{Small Object Understanding capability of Aerial scenarios}. In the A-VisDrone sub-dataset, the best performance of the model is GPT-5.2~\cite{gpt-5}, and it reaches 51.98\%. The second-best performance of the model is InternVL3\_5-14B~\cite{Internvl3_5}, and it exhibits 51.43\% performance. Similarly, in the other two sub-datasets, the best and second-best models in A-AITOD
are InternVL3\_5-14B~\cite{Internvl3_5} (71.87\%) and Qwen3-VL-8B-Instruct~\cite{qwen3} (70.53\%). The best and second-best models in A-AITOD are Gemini-3-flash-preview-nothinking~\cite{gemini} (61.89\%) and GPT-5.2~\cite{gpt-5} (55.90\%), respectively. (2)\textbf{The Performance Difference Analysis}. We find that even within the same Aerial scenario, MLLMs exhibit a significant performance difference. We think the performance gap across these benchmark datasets is primarily driven by scenario entropy and object density. Specifically, the AI-TOD dataset has relatively consistent backgrounds and allows the model to better distinguish tiny features, which leads to the highest accuracy. In contrast, VisDrone poses the greatest challenge due to severe occlusions and cluttered urban environments, which significantly increase the difficulty of feature extraction for MLLMs. (3) \textbf{Compared with Human Performance.} In three different datasets, the best performance of MLLM lagged behind human performance by 9.69\% (A-VisDrone), 21.88\% (A-AITOD), and 21.44\% (A-SODA), respectively. This indicates weak small-object recognition capabilities in Aerial scenarios, with the gap reaching as high as 21.88\%. This performance gap reveals that our current MLLMs still have a long way to go in achieving truly small object understanding.

\begin{table*}[h!]   
\begin{center}   
\caption{The more evaluation results on sub-datasets of the Aerial scenario. A-VisDrone, A-AITOD, and A-SODA denote the three created sub-datasets based on the original VisDrone, AITOD, and SODA, respectively. \sethlcolor{red!20}\hl{CE}, \sethlcolor{red!20}\hl{OC} denotes the two sub-tasks of foundational Perception. \sethlcolor{blue!20}\hl{CR}, \sethlcolor{blue!20}\hl{OL} denotes the two sub-tasks of Spatial Reasoning. \sethlcolor{green!20}\hl{SCC}, \sethlcolor{green!20}\hl{POI} denotes the two sub-tasks of Fine-Grained Understanding. $^\dagger$ denotes the active parameters. \textbf{\underline{Bold}} and \underline{underlined} denotes the best and second-best model performance, respectively.}
\renewcommand\arraystretch{0.9} 
\resizebox{\textwidth}{!}
{

\begin{tabular}{l|c|c|cccccc|c}   \hline  
Model & Scale & Dataset & \cellcolor{red!20} CE & \cellcolor{red!20} OC & \cellcolor{blue!20} CR & \cellcolor{blue!20} OL & \cellcolor{green!20} ~~SCC & \cellcolor{green!20} POI &  Average \\ \hline
\rowcolor{blue!1}
Random Chance  & - &  & 25.00 & 25.00 & 25.00 & 25.00 & 25.00 & 25.00 & 25.00  \\ 
\rowcolor{blue!1}
Human Performance  & - &  & 30.00 & 50.00 &  70.00 & 100 & 50.00 & 70.00 & 61.67  \\ 
\rowcolor{blue!1}
Gemini-3-flash-preview-nothinking~\cite{gemini}     &  -  &  &  41.25 &    31.71 &    48.25 &    34.44 &    39.88 &    \textbf{\underline{58.17}} & 42.28 \\
\rowcolor{blue!1}
GPT-5.2~\cite{gpt-5}     &  -  &  &  33.66 &    33.85 &    \textbf{\underline{55.25}} &    72.57 &    61.87 &    \underline{54.67}   & \textbf{\underline{51.98}} \\
\rowcolor{blue!1}
Grok-4-1-fast-no-reasoning~\cite{grok-4.1}     & - &  &  24.32 &    22.37 & 37.16 &    33.66 &    50.58 &    32.68  & 33.46  \\  
\rowcolor{blue!1}
Deepseek-VL2-tiny~\cite{deepseekvl2}& 3B/1B$^\dagger$ & &  31.71 &    24.12 &    42.61 &    25.10 &    18.48 &    36.96 & 29.83 \\
\rowcolor{blue!1}
InternVL3\_5-8B~\cite{Internvl3_5} & 8B & &  44.75 &    32.49 &    44.55 &    55.84 &    \underline{62.26} &    48.05 & 47.99 \\
\rowcolor{blue!1}
InternVL3\_5-14B~\cite{Internvl3_5} & 14B & &   45.72 &    37.35 &    52.33 &    66.15 &    61.28 &    45.72 & \underline{51.43} \\
\rowcolor{blue!1}
Kimi-VL-A3B-Instruct~\cite{Kimi-VL} &16B/3B$^\dagger$ & A-VisDrone &   40.27 &    \textbf{\underline{38.52}} &    49.81 &    50.00 &    43.58 &    49.81 & 45.33 \\
\rowcolor{blue!1}
MiniCPM-V-4\_5~\cite{MiniCPM-V4.5} & 8B & &    32.10 &    32.49 &    48.25 &    67.70 &    59.92 &    43.77 & 47.37\\
\rowcolor{blue!1}
QianFan-VL-8B~\cite{Qianfan-vl} & 8B & &  39.30 &    33.66 &    46.11 &    65.18 &    \textbf{\underline{63.04}} &    52.92 & 50.04 \\
\rowcolor{blue!1}
Qwen2.5-VL-3B-Instruct~\cite{Qwen2.5-VL} & 3B &  &  \textbf{\underline{47.47}} &    36.58 &    50.97 &    \textbf{\underline{80.35}} &    49.22 &    43.39 & 51.33 \\
\rowcolor{blue!1}
Qwen3-VL-4B-Instruct~\cite{qwen3} & 4B & &  38.33 &    37.16 &    44.55 &    60.12 &    \underline{62.26} &    48.64 & 48.51 \\ 
\rowcolor{blue!1}
Qwen3-VL-8B-Instruct~\cite{qwen3} & 8B & &  37.94 &    \underline{38.13} &    48.83 & 63.62 &    59.53 &    51.75 & 49.97  \\
\rowcolor{blue!1}
VLM-R1~\cite{VLM-R1} & 3B & &  \underline{46.50} &    35.99 &    \underline{53.31} &    \underline{74.90} &    49.42 &    42.61 & 50.46 \\
\rowcolor{blue!1}
Gemma3-4B~\cite{team2025gemma}  & 4B & &  36.58 & 29.38 &    35.41 &    23.74 & 47.47 & 40.66 & 35.54 \\
\rowcolor{blue!1}
Deepseek\_VL\_7B~\cite{lu2024deepseekvl}& 7B & &39.69 &    25.68 &    42.02 &    23.35 &    34.05 &    46.69 & 35.25 \\ \hline 

\rowcolor{blue!1}
Random Chance  & - &  & 25.00 & 25.00 &  25.00 & 25.00 & 25.00 & 25.00 & 25.00  \\ 
\rowcolor{blue!1}
Human Performance  & - &  & 87.50 & 87.50 &  100 & 100 & 87.50 & 100 & 93.75 \\ 
\rowcolor{blue!1}
Gemini-3-flash-preview-nothinking~\cite{gemini}     &  -  &  &  \textbf{\underline{81.95}} &    40.00 &    87.80 &    42.20 &    \underline{50.73} &   \textbf{\underline{89.27}}  & 65.33 \\
\rowcolor{blue!1}
GPT-5.2~\cite{gpt-5}     &  -  &  &  47.80 &    36.83 &    89.27 &    73.66 &    46.10 &    \underline{89.02}  & 63.78 \\
\rowcolor{blue!1}
Grok-4-1-fast-no-reasoning~\cite{grok-4.1}     & - &  &  28.78 &    25.37 &    74.63 &    42.20 &    30.24 &    67.07 & 44.72 \\  
\rowcolor{blue!1}
Deepseek-VL2-tiny~\cite{deepseekvl2}& 3B/1B$^\dagger$ & &     2.20 &    22.68 &    59.76 &    17.07 &    24.88 &    39.76 & 27.73\\
\rowcolor{blue!1}
InternVL3\_5-8B~\cite{Internvl3_5} & 8B & &    35.37 &    \underline{44.39} &    87.32 &    78.54 &    \underline{50.73} &    78.29 & 62.44 \\
\rowcolor{blue!1}
InternVL3\_5-14B~\cite{Internvl3_5} & 14B & &  71.95 &    \textbf{\underline{48.29}} &    90.73 &    72.20 &    \textbf{\underline{60.49}} &    87.56 & \textbf{\underline{71.87}} \\
\rowcolor{blue!1}
Kimi-VL-A3B-Instruct~\cite{Kimi-VL} &16B/3B$^\dagger$ & A-AITOD &   29.27 &    36.34 &    78.78 &    62.93 &    47.80 &    85.61 & 56.79 \\
\rowcolor{blue!1}
MiniCPM-V-4\_5~\cite{MiniCPM-V4.5} & 8B & &  39.02 &    24.15 &    88.29 &    \underline{86.34} &    40.24 &    81.71 & 59.96\\
\rowcolor{blue!1}
QianFan-VL-8B~\cite{Qianfan-vl} & 8B& &    12.93 &    38.78 &    \underline{91.22} &    \textbf{\underline{87.07}} &    48.78 &    79.76 & 59.76 \\
\rowcolor{blue!1}
Qwen2.5-VL-3B-Instruct~\cite{Qwen2.5-VL} & 3B &  &   52.44 &    30.73 &    83.66 &    49.27 &    45.61 &    71.46 & 55.53 \\
\rowcolor{blue!1}
Qwen3-VL-4B-Instruct~\cite{qwen3} & 4B & &   40.73 &    36.34 &    85.37 &    83.17 &    49.02 &    79.51 & 62.36 \\
\rowcolor{blue!1}
Qwen3-VL-8B-Instruct~\cite{qwen3} & 8B & &  \underline{73.41} &    38.05 &    \textbf{\underline{93.41}} & 80.00 &    49.51 &    88.78 & \underline{70.53} \\
\rowcolor{blue!1}
VLM-R1~\cite{VLM-R1} & 3B & & 50.00 &    30.24 &    80.73 &    57.56 &    45.61 &    71.95 & 56.02 \\
\rowcolor{blue!1}
Gemma3-4B~\cite{team2025gemma}  & 4B & &  8.78 & 29.51 &    81.71 & 18.05 & 34.15 & 83.17 & 42.56 \\
\rowcolor{blue!1}
Deepseek\_VL\_7B~\cite{lu2024deepseekvl}& 7B & &  1.95 &    22.93 &    82.20 &    22.20 &    28.05 &    75.12 & 38.74\\ \hline 

\rowcolor{blue!1}
Random Chance  & - &  & 25.00 & 25.00 & 25.00 & 25.00 & 25.00 & 25.00 & 25.00  \\ 
\rowcolor{blue!1}
Human Performance  & - &  & 71.43 & 57.14 &  100 & 100 & 85.71 & 85.71 & 83.33  \\ 
\rowcolor{blue!1}
Gemini-3-flash-preview-nothinking~\cite{gemini}     &  -  &  &  \textbf{\underline{57.32}} &    \textbf{\underline{46.65}} &    74.09 &    \textbf{\underline{59.15}} &    \textbf{\underline{55.79}} &    \textbf{78.35} & \textbf{\underline{61.89}} \\
\rowcolor{blue!1}
GPT-5.2~\cite{gpt-5}     &  -  &  &  44.21 &    37.80 &    \underline{74.70} &    \underline{54.57} &    \underline{51.22} &    \underline{72.87}  & \underline{55.90} \\
\rowcolor{blue!1}
Grok-4-1-fast-no-reasoning~\cite{grok-4.1}     & - &  & 32.01 &    37.20 &    45.43 &    39.33 &    39.94 &    42.38 & 39.38  \\  
\rowcolor{blue!1}
Deepseek-VL2-tiny~\cite{deepseekvl2}& 3B/1B$^\dagger$ & &     7.01 &    24.39 &    53.05 &    31.71 &    20.12 &    45.43 & 30.29\\
\rowcolor{blue!1}
InternVL3\_5-8B~\cite{Internvl3_5} & 8B & &  39.94 &    \underline{41.16} &    67.38 &    37.50 &    47.26 &    65.85 & 49.85 \\
\rowcolor{blue!1}
InternVL3\_5-14B~\cite{Internvl3_5} & 14B & &  \underline{53.35} &    40.24 &    71.04 &    42.38 &    45.73 &    65.55 & 53.05 \\
\rowcolor{blue!1}
Kimi-VL-A3B-Instruct~\cite{Kimi-VL} & 16B/3B$^\dagger$& A-SODA &   22.87 &    32.62 &    70.43 &    26.52 &    42.68 &    62.50 & 42.94 \\
\rowcolor{blue!1}
MiniCPM-V-4\_5~\cite{MiniCPM-V4.5} & 8B & &  29.27 &    25.91 &    71.95 &    46.04 &    39.33 &    69.82 & 47.05 \\
\rowcolor{blue!1}
QianFan-VL-8B~\cite{Qianfan-vl} & 8B & &  29.57 &    39.33 &    61.89 &    27.13 &    48.78 &    56.10 & 43.80\\
\rowcolor{blue!1}
Qwen2.5-VL-3B-Instruct~\cite{Qwen2.5-VL} & 3B &  &    40.24 &    16.16 &    \textbf{\underline{75.91}} &    33.23 &    28.96 &    67.68 & 43.70 \\
\rowcolor{blue!1}
Qwen3-VL-4B-Instruct~\cite{qwen3} & 4B & &  46.65 &    32.62 &    68.90 & 32.32  &    43.90 &    66.16 & 48.43 \\
\rowcolor{blue!1}
Qwen3-VL-8B-Instruct~\cite{qwen3} & 8B & &  49.09 &    32.93 &    74.39 & 37.80 &    41.46 &    66.46 & 50.36 \\
\rowcolor{blue!1}
VLM-R1~\cite{VLM-R1} & 3B & & 40.85 &    21.34 &    71.65 &    35.37 &    28.66 &    61.89 & 43.29\\
\rowcolor{blue!1}
Gemma3-4B~\cite{team2025gemma}  & 4B & & 27.44 & 14.94 & 53.96 & 15.55 & 24.39 & 48.78 & 30.84 \\
\rowcolor{blue!1}
Deepseek\_VL\_7B~\cite{lu2024deepseekvl}& 7B & & 12.80 &    16.16 &    57.32 &    21.95 &    21.04 &    52.13 & 30.23 \\ \hline 

\end{tabular}
}
\label{BVCXBVCBXCVBXCVBXBXCVBXVCBVCB} 
\end{center}   
\end{table*}

\vspace{+15pt}
\noindent \textbf{\large D: ~More detailed setting of training} \par
\vspace{+15pt}

\noindent \textbf{Settings.}  
In our experiments, we set the latest Qwen3-VL-2B-Instruct~\cite{qwen3} model as the baseline model. All settings of training are shown in Tab.~\ref{jkbjkjkgfjhkdjfglkhjdfklgjklh}. Here, we set the image\_max\_pixels as $4,194,304$ (\textit{i.e.,} $2048 \times 2048$) due to the high resolution. The detailed resolution size of different sub-datasets can be found in the Tab.~\ref{mvkxbvnkcnbjkcvnbjnvcxjkbnxcjbnjcxn}. Besides, we adopt the Lora~\cite{Lora} method to SFT and set the lora\_rank to 8.  
Note that due to the distinct characteristics of datasets across different scenarios, \textit{e.g.}, small object categories, number of images, and original image resolution. Therefore, we set different training epochs for each dataset to ensure the model is sufficiently trained without overfitting, as shown in Fig.~\ref{ahgjkdfhgnxmblsjk}. \par

\begin{table}[htbp]
\centering
\caption{The detailed hyperparameter settings for Qwen3-VL-2B-Instruct fine-tuning.}
\renewcommand\arraystretch{0.82} 
\scalebox{0.83}
{
\begin{tabular}{ll}
\toprule
Name & Value \\
\midrule
\textcolor{gray}{Model \& Input Architecture} & \\
base model & Qwen3-VL-2B-Instruct \\
image\_max\_pixels & $4,194,304$ ($2048 \times 2048$) \\
\midrule
\textcolor{gray}{Fine-tuning Method (LoRA)} & \\
stage & sft \\
finetuning\_type & lora \\
lora\_rank ($r$) & 8 \\
lora\_target & all  \\
\midrule
\textcolor{gray}{Optimization Hyperparameters} & \\
preprocessing\_num\_workers & 16 \\
lr\_scheduler\_type & cosine \\
warmup\_ratio & 0.1 \\
num\_train\_epochs\_Driving & 9.0 \\
num\_train\_epochs\_Aerial (A-VisDrone) & 6.0 \\
num\_train\_epochs\_Aerial (A-AITOD) & 4.0 \\
num\_train\_epochs\_Aerial (A-SODA) & 3.0 \\
num\_train\_epochs\_Underwater & 3.0 \\
per\_device\_train\_batch\_size & 1 \\
gradient\_accumulation\_steps & 8 \\
ddp\_timeout & 180000000 \\
\midrule
\textcolor{gray}{Hardware \& Precision} & \\
bf16 & true \\
preprocessing\_num\_workers & 16 \\
dataloader\_num\_workers & 4 \\
\bottomrule
\end{tabular}
}
\label{jkbjkjkgfjhkdjfglkhjdfklgjklh}
\end{table}

\noindent \textbf{Supervised Fine-Tuning Dataset Template.} In our paper, we also design a program for automatically generating training datasets. Note that this program is similar to that for generating the validation dataset, differing only in the output Template. The output Template is meticulously designed. Based on the automatic program, we generate different Supervised Fine-Tuning Datasets for all scenarios and datasets. Finally, we construct SOU-Train, a comprehensive supervised fine-tuning training dataset, with \textbf{11,226} rich and fine-grained VQA annotations. Here, we take three  Supervised Fine-Tuning Dataset Template cross different scenarios as examples, as shown below. Supervised Fine-Tuning Dataset Template 1, 2, and 3 denotes the Driving, Aerial, and Underwater scenarios.

\begin{cmt2}{}{}
\label{vcbmncvmblnjfdkpogkiposfok}
$\left \langle \textbf{user} \right \rangle$  Based on the visual information in the image, please describe any small objects detail in the scene. \par
$\left \langle \textbf{assistant} \right \rangle$ The image describes an autonomous driving scenario for small object understanding tasks.The image contains several small object categories, including: people, traffic-sign. There are 2 different categories in total, with a cumulative count of 4 small objects.Based on the spatial analysis of all small objects in the image, the category of people possesses the smallest ordinate value  among all bounding box coordinates of the small object. Therefore, it is logically determined that the topmost small object belongs to the category of people in this image.
\end{cmt2}

\begin{cmt2}{}{}
\label{vcbmncvmblnjfdkpogkiposfok}
$\left \langle \textbf{user} \right \rangle$  Based on the visual information in the image, please describe any small objects detail in the scene. \par
$\left \langle \textbf{assistant} \right \rangle$ The image describes an aerial scenario for small object understanding tasks.The image contains several small object categories, including: airplane, vehicle. There are 2 different categories in total, with a cumulative count of 14 small objects.Based on the spatial analysis of all small objects in the image, the category of airplane possesses the smallest ordinate value among all bounding box coordinates of the small object. Therefore, it is logically determined that the topmost small object belongs to the category of airplane in this image.
\end{cmt2}

\begin{cmt2}{}{}
\label{vcbmncvmblnjfdkpogkiposfok}
$\left \langle \textbf{user} \right \rangle$  Based on the visual information in the image, please describe any small objects detail in the scene. \par
$\left \langle \textbf{assistant} \right \rangle$ The image describes an under water scenario for small object understanding tasks.The image contains several small object categories, including: fish. There are 1 different categories in total, with a cumulative count of 2 small objects.Based on the spatial analysis of all small objects in the image, the category of fish possesses the smallest abscissa value among all bounding box coordinates of the small object. Therefore, it is logically determined that the leftmost small object belongs to the category of fish in this image.
\end{cmt2}


\vspace{+15pt}
\noindent \textbf{\large E: ~More results of different model scales} \par
\vspace{+15pt}

\begin{figure*}[htbp]
	\centering
	\begin{minipage}{1\linewidth}
		\centering
		\includegraphics[width=0.96\linewidth]{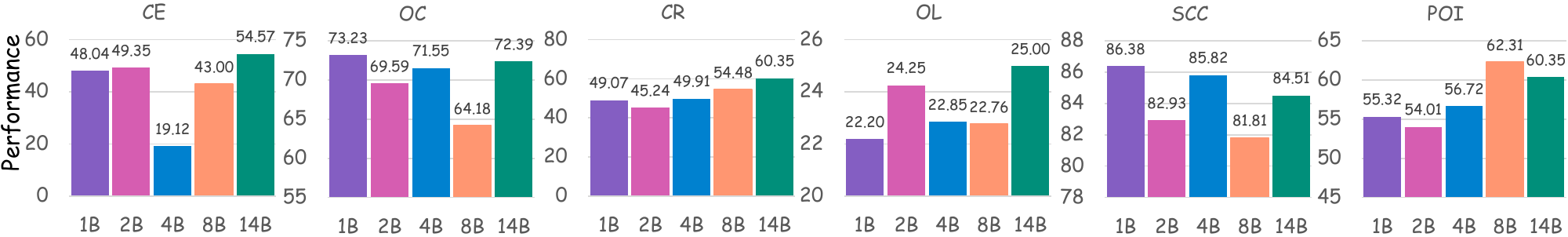}
		\subcaption{Driving}
	\end{minipage}
	\qquad
	\begin{minipage}{1\linewidth}
		\centering
		\includegraphics[width=0.96\linewidth]{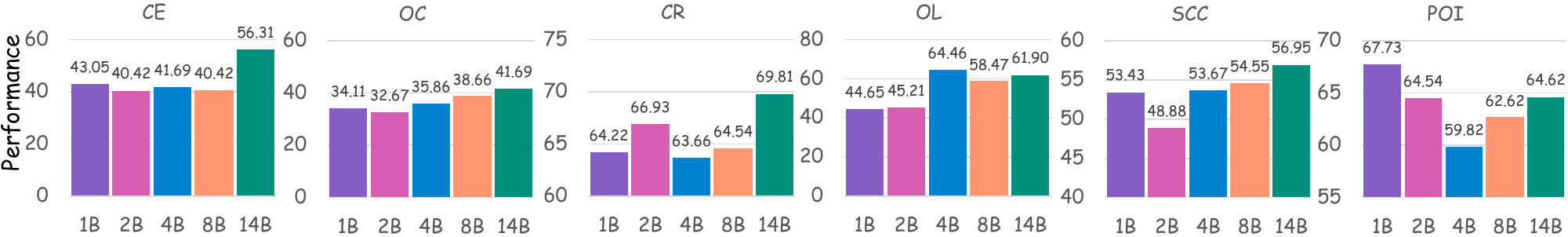}
		\subcaption{Aerial (three sub-tasks)}
	\end{minipage}
    	\qquad
	\begin{minipage}{1\linewidth}
		\centering
		\includegraphics[width=0.96\linewidth]{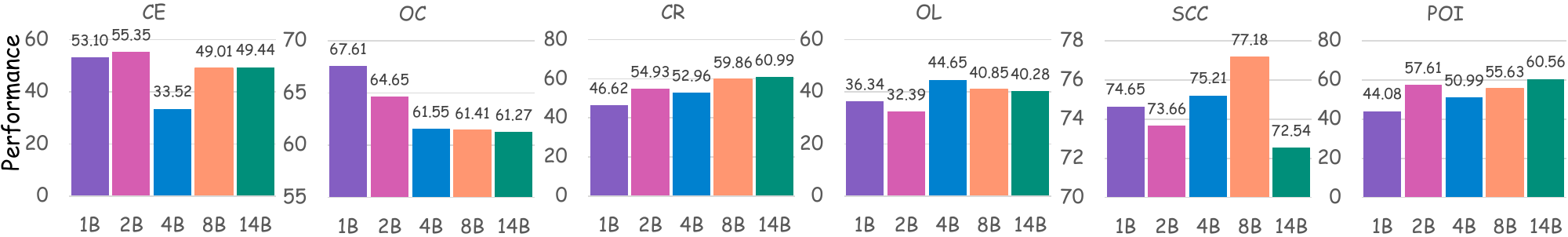}
		\subcaption{Underwater}
	\end{minipage}
    	\qquad
	\begin{minipage}{1\linewidth}
		\centering
		\includegraphics[width=0.96\linewidth]{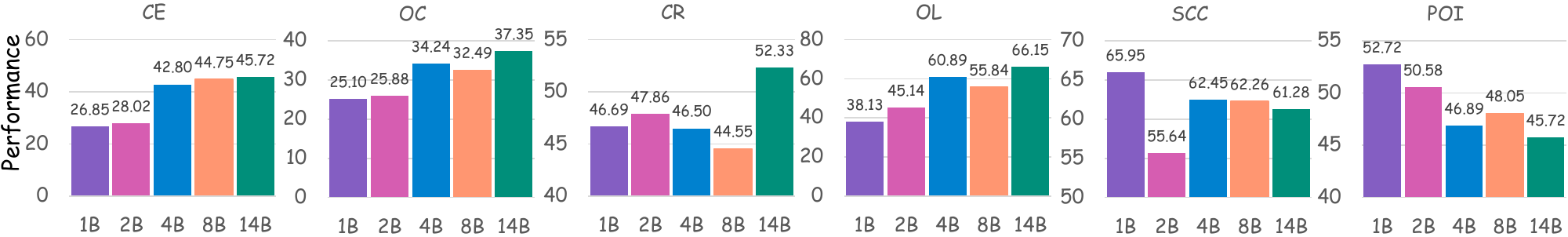}
		\subcaption{Aerial (A-VisDrone)}
	\end{minipage}
    	\qquad
	\begin{minipage}{1\linewidth}
		\centering
		\includegraphics[width=0.96\linewidth]{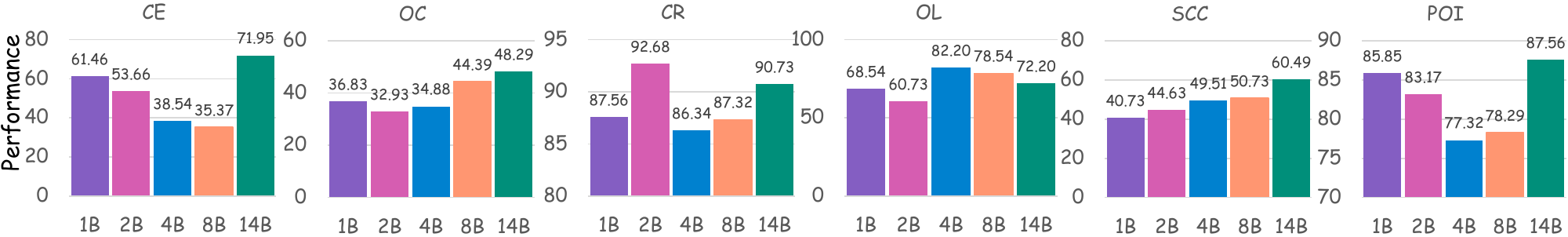}
		\subcaption{Aerial (A-AITOD)}
	\end{minipage}
    	\qquad
	\begin{minipage}{1\linewidth}
		\centering
		\includegraphics[width=0.96\linewidth]{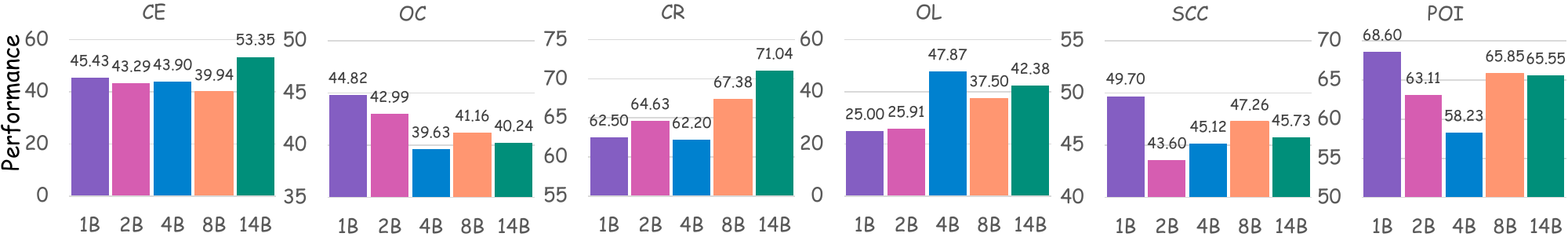}
		\subcaption{Aerial (A-SODA)}
	\end{minipage}
\caption{The more detailed results of different model scales. Here, we report more results in different scenarios and sub-datasets. We reveal the shortcomings of current MLLMs across different sub-tasks and suggest some directions for future improvements, \textit{e.g.}, in the Aerial (A-VisDrone) scenario, as the model scale increases, the performance of the model in POI (Peripheral Object Identification)is decreased. These results suggest that increasing model scale does not inherently resolve the bottleneck in fine-grained understanding ability of small objects based on their relative spatial position.}
\label{nmcvxnblkojhkgfjdklhjgfdhopgf}
\end{figure*}

\noindent In this section, we present more detailed results for different model scales, including three scenarios and the performance of various sub-tasks in the proposed SOU task, as shown in Fig.~\ref{nmcvxnblkojhkgfjdklhjgfdhopgf}. We can observe that, across different scenarios, despite some fluctuations, most sub-tasks follow a consistent pattern: 
In most cases, optimal sub-task performance is achieved at 14B (larger model sizes generally yield better sub-task performance). In the CE (Category Enumeration) sub-task of the Aerial (A-VisDrone) scenario, performance evolves from 26.85\%$\rightarrow$28.02\%$\rightarrow$42.80\%$\rightarrow$44.75\%$\rightarrow$45.72\%. Similarly, in the SCC (Specific Category Counting) sub-task of the Aerial (A-AITOD) scenario, performance progresses from 40.73\%$\rightarrow$ 44.63\%$\rightarrow$49.51\%$\rightarrow$50.73\%$\rightarrow$60.49\%. (2) Besides, we also observe the opposite phenomenon in certain sub-tasks: MLLMs exhibit declining performance. For instance, in the OC (Object Counting) sub-task within the Underwater scenario, as model size increases, model performance changes from 67.61\%$\rightarrow$64.65\%$\rightarrow$61.55\%$\rightarrow$61.41\%$\rightarrow$61.27\%. In the POI (Peripheral Object Identification) task within the Aerial (A-VisDrone) scenario, model performance deteriorated from 52.72\%$\rightarrow$50.58\%$\rightarrow$46.89\%$\rightarrow$48.05\%$\rightarrow$45.72\%. According to these performances across different sub-tasks, we observe that MLLMs exhibit gaps in certain capabilities across various scenarios, \textit{e.g.,} Category Enumeration and Peripheral Object Identification. These findings represent a key direction for future improvements to MLLMs in small object understanding tasks.

\end{document}